\documentclass[accepted]{uai2025} % after acceptance, for a revised version

\usepackage[american]{babel}

\usepackage{natbib} % has a nice set of citation styles and commands
\usepackage[utf8]{inputenc}
\usepackage[T1]{fontenc}
\usepackage{amsfonts}
\usepackage{nicefrac} 
\usepackage{microtype}
\usepackage{graphicx}
\usepackage{amsmath}
\usepackage{amsthm}
\usepackage{amssymb}
\usepackage{amsfonts}
\usepackage{upgreek}
\usepackage{mathtools}
\usepackage{booktabs}
\usepackage{bbm}
\usepackage{bm}
\usepackage{multirow}
\usepackage[dvipsnames]{xcolor}
\usepackage{subcaption}
\usepackage{enumitem}
\usepackage{url}
\usepackage{upgreek}
\def\eqref#1{equation~\ref{#1}}
\def\1{\bm{1}}

\def\rv{{\textnormal{v}}}

\def\rz{{\textnormal{z}}}

\def\rvu{{\mathbf{i}}}

\def\rvu{{\mathbf{u}}}

\def\rvx{{\mathbf{x}}}
\def\rvy{{\mathbf{y}}}

\def\vx{{\bm{x}}}
\def\vy{{\bm{y}}}

\DeclareMathAlphabet{\mathsfit}{\encodingdefault}{\sfdefault}{m}{sl}
\SetMathAlphabet{\mathsfit}{bold}{\encodingdefault}{\sfdefault}{bx}{n}

\def\gD{{\mathcal{D}}}

\def\gF{{\mathcal{F}}}

\def\gH{{\mathcal{H}}}

\def\gL{{\mathcal{L}}}

\def\gO{{\mathcal{O}}}

\def\gR{{\mathcal{R}}}
\def\gS{{\mathcal{S}}}
\def\gT{{\mathcal{T}}}

\def\gX{{\mathcal{X}}}
\def\gY{{\mathcal{Y}}}

\newcommand{\gPsi}{\mathnormal{\Psi}}

\DeclareMathOperator*{\argmax}{arg\,max}

\hypersetup{colorlinks=true, allcolors=blue}

\newcommand{\ie}{\emph{i.e.}}
\newcommand{\eg}{\emph{e.g.}}

\theoremstyle{plain}
\newtheorem{theorem}{Theorem}[section]
\newtheorem{proposition}[theorem]{Proposition}
\newtheorem{lemma}[theorem]{Lemma}

\newtheorem{definition}[theorem]{Definition}

\title{On Continuous Monitoring of Risk Violations under Unknown Shift}

\author[1]{Alexander Timans\thanks{Equal contributions. $\quad$ Corresponding authors: $<$\texttt{a.r.timans@uva.nl}, \texttt{r.verma@uva.nl}$>$}$^,$}
\author[1]{Rajeev Verma$^{*,}$}
\author[2]{Eric Nalisnick}
\author[1]{Christian A. Naesseth}

\affil[1]{%
    UvA-Bosch Delta Lab, University of Amsterdam
}
\affil[2]{%
    Department of Computer Science, Johns Hopkins University
  }
\begin{document}
\maketitle

%%%%%%%%%%%%%%%%%%%%%%%%%%%%%%%%
% TEXT
%%%%%%%%%%%%%%%%%%%%%%%%%%%%%%%%

\begin{abstract}
    \looseness=-1Machine learning systems deployed in the real world must operate under dynamic and often unpredictable distribution shifts. This challenges the validity of statistical safety assurances on the system's risk established beforehand. Common risk control frameworks rely on fixed assumptions and lack mechanisms to continuously monitor deployment reliability. In this work, we propose a general framework for the real-time monitoring of risk violations in evolving data streams. Leveraging the `testing by betting' paradigm, we propose a sequential hypothesis testing procedure to detect violations of bounded risks associated with the model's decision-making mechanism, while ensuring control on the false alarm rate. Our method operates under minimal assumptions on the nature of encountered shifts, rendering it broadly applicable. We illustrate the effectiveness of our approach by monitoring risks in outlier detection and set prediction under a variety of shifts.
\end{abstract}

\section{Introduction}
\label{sec:intro}

\begin{figure*}[!t]
    \centering
    \includegraphics[width=0.95\linewidth]{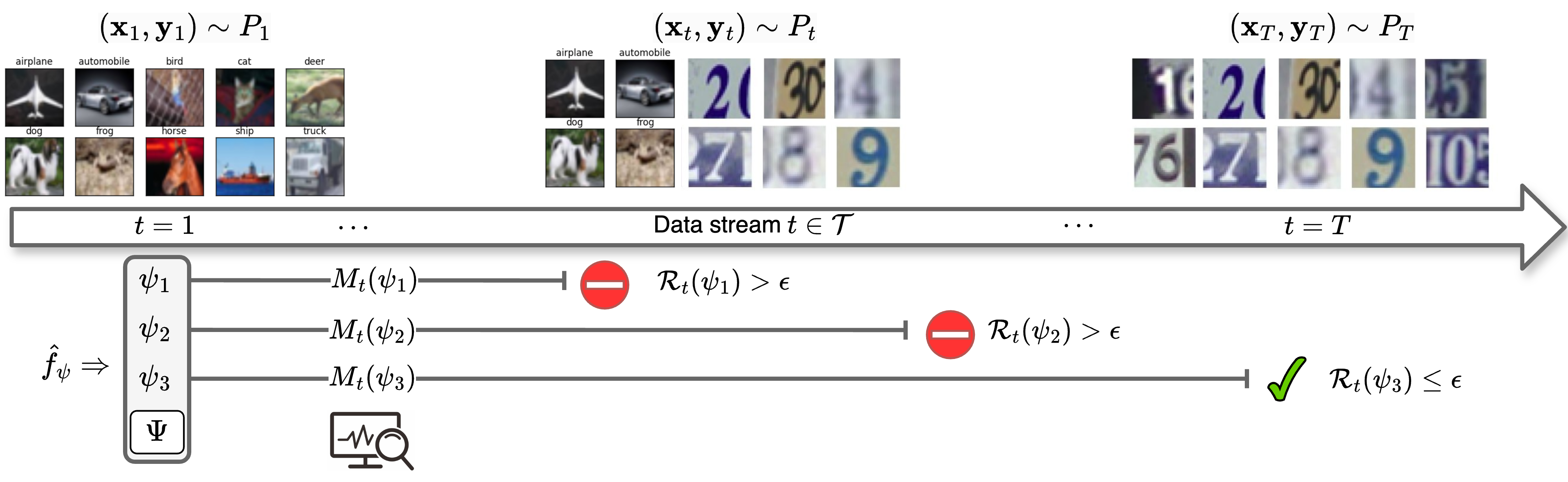}
    \caption{
    We consider an evolving data stream $t = 1, \dots, T$ susceptible to distribution shifts, \ie~observations are drawn from a time-dependent distribution $P_t$ at each step. A predictor $\hat{f}_{\psi}$ is equipped with a decision-making mechanism governed by a threshold $\psi$ (\eg~on outlier flagging). At deployment, we monitor each candidate $\psi \in \gPsi$ using a sequential testing process $M_t(\psi)$ which collects evidence for or against risk violations. Risk-violating thresholds are then marked as unreliable.
    }
\label{fig:infographic}
\end{figure*}

\looseness=-1The increasing demand for reliable predictions from machine learning systems has driven the development of statistical frameworks for \emph{distribution-free risk control} \citep{angelopoulos2021learn, bates2021distribution}. Such frameworks rely on data-driven inference to achieve their goal, leveraging representative held-out data to determine suitable parameters guiding an application-specific risk, \eg~selecting a threshold value for outlier flagging. The hope is that the user can then employ the determined settings indefinitely to aid in reliable decision-making. However, the common validation versus deployment mismatch in machine learning systems has the potential to thwart any `quality assurance' stamp these methods derive from their static inference. Challenges like outliers, distribution shifts and feedback loops are commonplace \citep{koh2021wilds}. In fact, \citet{van2023accurate} argue that an effective machine learning model should \emph{actively} affect the real-world---distribution shift is then not merely an artifact or deployment challenge, but rather a manifestation of a successfully operating system. Hence, any decision-making parameters necessitate \emph{continuous monitoring} during deployment, and the user should be notified when statistical reliability is faltering.

\looseness=-1We address this problem by proposing a general framework for the real-time continuous monitoring of bounded risks in evolving data streams, and raising a signal when desired risk levels are in danger of violation. Since alarm signals may trigger costly preventive measures, \eg~a production line stop in manufacturing or default loan denial in credit underwriting, it is crucial that false alarms are not raised too often, and our approach effectively controls this rate. We explicitly limit any assumptions on the deployment setting or nature of encountered data, rendering operability under arbitrary or \emph{unknown shifts}. To achieve our goal we adopt the `testing by betting' paradigm \citep{ramdas2023game}, and cast our monitoring task as a sequential hypothesis testing problem. Leveraging the framework's natural error control properties, our resulting monitoring procedure remains both efficient and statistically rigorous. To summarize, our contributions include:

\begin{itemize}[leftmargin=13pt, itemsep=0pt, topsep=0pt]
    \item  In \autoref{sec:method}, we motivate sequential testing as a natural approach to continuous risk monitoring, place it in the context of `testing by betting' and re-interpret the prior method of \citet{Feldman2022AchievingRC} under this lens.
    \item In \autoref{sec:theory}, we theoretically outline the statistical properties of our approach, including control over the false alarm rate, asymptotic consistency, and, under some conditions, bounds on the detection time of violations (\hyperref[thm:detection-delay-argument]{Prop.~\ref{thm:detection-delay-argument}}).
    \item In \autoref{sec:exp}, we demonstrate the efficacy of our approach against baselines for risk monitoring in outlier detection (\autoref{subsec:exp-ood}) and set prediction tasks (\autoref{subsec:exp-sets}), employing real-world datasets and different shift scenarios including natural temporal shifts.
\end{itemize}

\section{Risk and Problem Formulation}
\label{sec:background}

\looseness=-1We next describe our notation, problem setting and task in detail, highlighting some key distinctions to existing work.

\paragraph{Notation and risk quantity.} \looseness=-1Let $\gX \times \gY$ denote the sample space with a data-generating distribution $P$ over it, and $\rvx, \rvy$ random variables with realizations $\vx, \vy$.\footnote{Upright lettering denotes random variables and italic lettering their realizations. Boldening denotes multi-dimensional quantities.} We consider access to the outputs of a base predictor $\hat{f}: \gX \rightarrow \gS$, where $\gS \subseteq \mathbb{R}^{|\gY|}$ for classification or $\gS \subseteq \mathbb{R}$ for regression. This model may have been explicitly trained by the user, but can in particular denote a pretrained model without internal access, \eg~accessible via an API. Next, similar to existing approaches for risk control \citep{angelopoulos2021learn, angelopoulos2024crc, Feldman2022AchievingRC}, we equip the model with a general decision-making mechanism of the form
\begin{equation}
\label{eq:def-threshold-pred}
    \hat{f}_{\psi}(\rvx) = 
    g(\hat{f}(\rvx), \psi),
\end{equation}
where $\psi \in \gPsi,\,\gPsi \subseteq [0,1]$ denotes a particular threshold value and $g$ a generic operator instantiated for each task-specific thresholding mechanism. For example, we can define $g$ as a binary decision on outlier flagging given some outlier score computed using $\hat{f}$ (see \autoref{subsec:exp-ood}). Finally, a notion of error for $\hat{f}_{\psi}$ and any particular threshold $\psi$ is captured by a problem-specific \emph{supervised and bounded} loss function $\ell: \gX \times \gY \times \gPsi \rightarrow \gL,\,\gL \subseteq [0,1]$, and the resulting \emph{true population risk} is given by the expected loss
\begin{equation}
\label{eq:def-risk}
    \gR(\psi) = \mathbb{E}_{P}[\ell(\hat{f}_{\psi}(\rvx), \rvy)].
\end{equation}
Because $\ell \in \gL$ is bounded it also follows that $\gR(\psi) \in [0,1]$. Boundedness of the loss constitutes our key restriction, but we place no conditions on the particular distribution of losses within those bounds \citep{waudby2024estimating}. To simplify notation we additionally define $\rz = \ell(\hat{f}_{\psi}(\rvx), \rvy)$ as a random variable of the loss with realization $z$, and equivalently express the risk in \autoref{eq:def-risk} as $\gR(\psi) = \mathbb{E}_{P}[\rz]$. Crucially, $\gR(\psi)$ denotes the quantity of interest for which safety assurances of some form are desired in order to robustify decisions made using $\hat{f}_{\psi}$ (and indirectly, $\hat{f}$).

\paragraph{Static risk control.} \looseness=-1Assume the deployment of $\hat{f}_{\psi}$ on new \emph{i.i.d.} test data $\gD_{test} \sim P_0$, and access to representative labelled \emph{i.i.d.} calibration data $\gD_{cal} \sim P_0$. Following existing frameworks of risk control such as \emph{RCPS} \citep{bates2021distribution} or \emph{Learn-then-Test} \citep{angelopoulos2021learn}, $\gD_{cal}$ can be leveraged to identify a subset $\hat{\gPsi} \subseteq \gPsi$ of \emph{risk-controlling} thresholds which ensures a high-probability upper bound on the population risk. That is, for any $\hat{\psi} \in \hat{\gPsi}$ we may state that $\mathbb{P}(\gR(\hat{\psi}) \leq \epsilon) \geq 1 - \delta$ holds. The risk level $\epsilon \in (0,1)$ and probability level $\delta \in (0,1)$ are user-specified, and dictate how tightly the risk is to be controlled. For instance, selecting low values for both $\epsilon$ and $\delta$ will enforce strong guarantees but may result in overly conservative decision-making on the basis of a chosen $\hat{\psi}$. Crucially, these approaches operate in a \emph{static} batch setting where the set $\hat{\gPsi}$ is computed once and deployed indefinitely, and are limited by their assumption on a \emph{static} distribution $P_0$ over time.

\paragraph{Data stream setting under shift.} \looseness=-1Instead, let us consider a more dynamic stream setting at deployment time. Given a time index set $\gT = \{1, \dots, T\}$, at every time step $t \in \gT$ a covariate $\vx_t$ is obtained, a decision is made using $\hat{f}_{\psi}(\vx_t)$, and subsequently $\vy_t$ is revealed and the loss $z_t$ measured. Thus, the flow of information at each step follows as \emph{covariate $\rightarrow$ decision $\rightarrow$ label $\rightarrow$ loss}, and at time $t$ the observational history $\{(\vx_i, \vy_i, z_i)\}_{i=1}^{t-1}$ is available. If we assume that the test stream originates \emph{i.i.d} $(\vx_t, \vy_t)_{t \in \gT} \sim P_0$, risk control frameworks as above could be directly applied after observing sufficient samples, and we elaborate further on this simpler setting in \autoref{subsec:connection-methods}. In this work, we address the challenging extension to the stream case under \emph{time-dependent and unknown} distribution shifts. Specifically, we consider a data stream observed as $(\vx_t, \vy_t) \sim P_t$ for $t \in \gT$, where samples at every time step originate from a time-dependent distribution $P_t$ which may shift, and in particular tends to deviate away from any initial $P_0$. Our risk quantity of interest then becomes $\gR_t(\psi) = \mathbb{E}_{P_t}[\rz_t]$, the \emph{time-dependent} true population risk at any given time $t$, and any obtained threshold set $\hat{\gPsi}_t \subseteq \gPsi$ is similarly time-dependent. We suppose minimal knowledge and place \emph{no assumptions} on the nature of the shift, which may be caused by a single static jump, gradual, \emph{etc.}, and originate in the covariates, labels, or both. Expectedly, the resulting high unpredictability on future risk development renders it substantially harder to provide safety assurances of any kind, but poses a commonly encountered problem setting in practice. Faced with such a challenge at deployment, we examine how to continuously monitor the true risk $\gR_t(\psi)$ for candidates $\psi$ and identify when violations of the form $\gR_t(\psi) > \epsilon$ occur. 

\section{Risk Monitoring as Sequential Testing By Betting}
\label{sec:method}

\looseness=-1We next outline our approach to risk monitoring leveraging sequential hypothesis testing. We motivate how such a testing framework naturally arises by recasting our stream setting as a forecasting `game' between two agents---the forecaster and nature---and formalizing the collected evidence as an error accumulation process. The procedure is then placed in the context of sequential `testing by betting' \citep{ramdas2023game}, thereby enjoying the practicality as well as the rigour of the framework. Finally, we theoretically connect our approach to related methods in \autoref{subsec:connection-methods}.

\paragraph{A sequential forecasting game.} \looseness=-1Consider a game between two agents, the \emph{forecaster} and \emph{nature} (\ie~the environment). The forecaster provides a guess $\pi_t$ for the true risk at time $t$ given their knowledge of the observational history, formally encapsulated in the filtration $\mathcal{F}_{t-1} = \sigma(\{(z_1, \pi_1), (z_2, \pi_2), \dots, (z_{t-1}, \pi_{t-1})\})$ (refer \autoref{app:math-defn.-and-terminology} for technical definitions). Should the forecaster desire to minimize the mean squared prediction error ${\mathbb{E}_{P_t}[(\rz_t -  \pi_t)^{2}  \ \vert \ \mathcal{F}_{t-1}]}$, thei best guess is given by $\pi_t = \mathbb{E}_{P_t}[\rz_{t} \ \vert \ \mathcal{F}_{t-1}]$. Nature then reveals the value of $\rz_t$, leading to an observable \emph{discrepancy} $\delta_t = z_t - \pi_t$ representing the incurred forecasting error. As the game is repeated, a sequence of discrepancies $(\delta_t)_{t \in \mathcal{T}}$ is iteratively built. Crucially, if the forecaster continues to make their best guess at every step, the resulting discrepancy process forms a martingale difference sequence, \ie, $\mathbb{E}_{P_t}[\updelta_{t} \ \vert \ \mathcal{F}_{t-1}] = 0$ and hence asymptotically $\frac{1}{t}\sum_{i=1}^{t}\delta_i \rightarrow 0$ as $t \rightarrow \infty$. Thus, under the forecaster's best strategy asymptotic alignment between forecasts and actual outcomes is ensured, and systematic deviations in the discrepancies (\ie~error accumulation) can serve as evidence, or a testing signal, for such alignment.

\looseness=-1Adopting the game to our problem setting, assume the forecaster's guess is upper-bounded as $\mathbb{E}_{P_t}\left[\rz_t \ \vert \ \mathcal{F}_{t-1}\right] \leq \epsilon$. In general, nature has no obvious incentive to align its realizations of $\rz_t$ with the forecaster. However, in our setting the outcomes are directly affected by the choice of threshold $\psi$ since $\rz_t = \ell(\hat{f}_{\psi}(\rvx_t), \rvy_t)$, rendering the associated discrepancy process useful for testing. For each candidate $\psi \in \gPsi$, a formal hypothesis test on alignment at risk level $\epsilon$ can be formulated as
\begin{align}
\label{eq:hypotheses-formulation}
\begin{split}
    H_{0}(\psi) &: \mathbb{E}_{P_t}\left[\rz_{t} \ \vert \ \mathcal{F}_{t-1}\right]\leq\epsilon \; \forall t \in \gT \,\quad \text{(risk controlled)} \\
    \quad H_{1}(\psi) &: \exists t \in \gT: \mathbb{E}_{P_t}\left[\rz_{t} \ \vert \ \mathcal{F}_{t-1}\right] > \epsilon, \quad \text{(risk violated)}
\end{split}
\end{align}
and our game suggests that the threshold's discrepancy process $(\delta_{t})_{t \in \gT}$ can provide the necessary testing evidence. 

\paragraph{Example test statistic.} \looseness=-1Given the sequential test in \autoref{eq:hypotheses-formulation}, how should the discrepancy sequence be leveraged to construct a test statistic? A straightforward choice is the cumulative process $M_t(\psi) = \sum_{i=1}^{t} \lambda_i \cdot \updelta_i = \sum_{i=1}^{t} \lambda_i (\rz_i - \epsilon)$, where $\lambda_t$ denotes a non-negative weight associated with the `trust' placed in the aggregated evidence at time $t$, dictating how $M_t(\psi)$ evolves. Intuitively, if $H_0(\psi)$ is true and the risk is indeed bounded by $\epsilon$, then the forecaster's guesses should be well aligned and discrepancies exhibit little systematic effects. In that case, $M_t(\psi)$ forms a \emph{supermartingale}, meaning that it is not expected to increase since $
\mathbb{E}_{P_t} [\updelta_t \mid \mathcal{F}_{t-1}] \leq 0$. On the other hand, consistent evidence indicating the risk's growth beyond $\epsilon$ will accumulate and drive the growth of $M_t(\psi)$, signaling evidence for rejection in favour of $H_1(\psi)$. Thus the cumulative process provides an viable test statistic for \autoref{eq:hypotheses-formulation}, and we further expand on this approach in \autoref{app:sum-process}.

\paragraph{Test supermartingales and testing by betting.} \looseness=-1While the aforementioned summation statistic offers a valid testing procedure, it is not necessarily \emph{efficient} in the sense of optimally accumulating evidence. That is, we want to accumulate the necessary evidence as fast as possible should a risk violation occur. To that end, a rich body of literature on sequential testing through the lens of `testing by betting' can be leveraged \citep{ramdas2023game}. Specifically, rather than via summation we may consider the multiplicative accumulation of discrepancies as
\begin{align}
\label{eq:test-supermartingale}
    M_t\left(\psi\right) = \prod_{i=1}^{t}\left(1 + \lambda_{i}\cdot \updelta_{i}\right) = \prod_{i=1}^{t}\left(1 + \lambda_{i}\left(\rz_i - \epsilon\right)\right),
\end{align}
yielding a universal representation of a \emph{test supermartingale} if we ensure $M_0 = 1$ and $(\lambda_t)_{t \in \mathcal{T}}$ to be a \emph{predictable} process based only on past observations \citep{ramdas2024hypothesis}. That is, $\lambda_t$ may only depend on $\{z_i\}_{i=1}^{t-1}$ (and is thus measurable w.r.t. $\mathcal{F}_{t-1}$). A game-theoretic interpretation can be given to the sequential test and each component in \autoref{eq:test-supermartingale}\footnote{The evidence collection process $M_{t}(\psi)$ can be interchangeably referred to as a \emph{test (super)martingale} by its mathematical properties, \emph{wealth process} by its betting interpretation, or \emph{E-process} in the context of the sequential testing literature.}. The forecaster is actively betting against the null hypothesis starting from an initial wealth of $M_0 = 1$, and $M_t(\psi)$ describes the \emph{wealth process} at every subsequent betting round $t$. The betting rate $\lambda_t$ denotes the proportion of wealth gambled at each step, and $(z_t - \epsilon)$ the resulting pay-off once nature reveals $\rz_t$. Should $H_0(\psi)$ hold, then no betting strategy is expected to systematically increase wealth. On the other hand, a betting strategy resulting in meaningful wealth accumulation points towards evidence against the null. A rejection threshold can be employed to reach a final testing decision with \emph{stopping time} $\tau(\psi) \in \mathcal{T}$, denoting the time step at which $H_0(\psi)$ has been ruled out by the wealth process.

\paragraph{Constructing threshold confidence sets.} \looseness=-1Since the risk associated with every threshold needs to be monitored simultaneously, we instantiate a number of wealth processes $M_t(\psi)$ in parallel, one for each candidate $\psi \in \gPsi$. Their joint behaviour can be encapsulated in a \emph{confidence set} ($\psi$-CS) of valid thresholds at every time step $t$, constructed as
\begin{equation}
\label{eq:psi-cs}
    C_{t}^{\psi} = \{\psi \in \gPsi \ : \ M_{t}\left(\psi\right) < 1/\delta\}.
\end{equation}
That is, using the predefined risk control parameters $\epsilon, \delta$, the confidence set $C_{t}^{\psi}$ denotes the set of thresholds at time $t$ for which $H_0(\psi)$ has not yet been rejected. It is, in effect, the equivalent of the threshold set $\hat{\gPsi}_t \subseteq \gPsi$ described in \autoref{sec:background} using the particular rejection threshold $1/\delta$. Crucially, by leveraging the stopping rule $1/\delta$ and test martingale properties of $M_t(\psi)$, any threshold that does \emph{not} violate the risk level $\epsilon$ at time $t$ is guaranteed to be included in $C_{t}^{\psi}$ with high probability, \ie, it holds that $\mathbb{P}_{H_0}(\forall t \in \mathcal{T} \ : \ M_{t}(\psi) < 1/\delta) > 1-\delta$. We interpret this Type-I error control property as a \emph{false alarm guarantee} on erroneous rejection, and elaborate upon it in \autoref{sec:theory}. In addition, the size of the $\psi$-CS can be interpreted as an indicator for the stream's shift intensity, and thus the underlying model's deployment reliability. A constant set size indicates temporally stable threshold choices are available, whereas a shrinkage of $C_{t}^{\psi}$ towards zero implies that all thresholds eventually signal risk violation, necessitating a more substantial model update using the observational history. Since we are pre-occupied with risk \emph{monitoring} only, we leave the discussion on model updating, or \emph{safe adaptation}, for future work.

\paragraph{Practical considerations.} \looseness=-1An important distinction to stress is that any false alarm guarantee holds \emph{across time} for every threshold, and not \emph{across thresholds} at every time step. Thus no guarantees can be given on an adaptive strategy to select a particular $\hat{\psi}_t \in C_{t}^{\psi}$ at every step, unless multiple testing corrections (which we do not consider here) are introduced to control for the multi-stream setting, \eg~drawing inspiration from \cite{xu2024online, dandapanthula2025multiple}. Empirically, one may adopt strategies such as selecting a stable threshold that persists over extended time horizons or a threshold to maximize significant results (\eg~the minimum value). These choices also relate to the \emph{risk profile} of $\gR_t(\psi)$, which dictates if a `trivial' stable solution (that may be very conservative) is available, and facilitates the interpretability of the obtained $\psi$-CS. A preferable risk profile will behave both monotonically across time (\eg, $\lim_{t \rightarrow 0} \gR_t(\psi) = 0 \text{ and } \lim_{t \rightarrow T} \gR_t(\psi) = 1$) as well as across thresholds (\eg, $\lim_{\psi \rightarrow 0} \gR_t(\psi) = 0 \text{ and } \lim_{\psi \rightarrow 1} \gR_t(\psi) = 1$). However, we do not assume such conditions and our experiments in \autoref{sec:exp} address non-monotonic behaviour in either argument.

\subsection{Relation to Other Approaches} 
\label{subsec:connection-methods}
\looseness=-1We next draw connections to other notions of risk control in the literature. Most notably, we leverage our formulation in terms of discrepancy processes to provide a novel interpretation of rolling risk control \citep{Feldman2022AchievingRC} as an implicit, adaptive form of sequential testing with asymptotic guarantees (as opposed to finite-sample). We then contrast our shifting stream setting with the simpler \emph{i.i.d.} case.

\paragraph{Sequential testing and rolling risk control.} \looseness=-1Proposed by \cite{Feldman2022AchievingRC} as an extension of \cite{Gibbs2021AdaptiveCI} to bounded risks beyond the miscoverage rate, \emph{rolling risk control} (RRC) aims to track the running estimate $\frac{1}{t} \sum_{i=1}^{t} z_i$ and ensure its asymptotic adherence to the risk level $\epsilon$ via the update rule $\psi_t = \psi_0 + \sum_{i=1}^{t-1} \gamma \, (z_i - \epsilon)$, where $\gamma>0$ denotes a step size. The `calibration parameter' $\psi$ governs the behaviour of their set predictor, rendering it an instantiation of our threshold model $\hat{f}_{\psi}$ (see also \autoref{subsec:exp-sets}). Procedurally, the model is initialized with value $\psi_0$ and RRC incrementally updates the parameter at every time step following the rule. Leveraging our discrepancy process interpretation, we can directly observe that RRC accumulates evidence via discrepancies $\delta_t = z_t - \epsilon$ over time, and is mathematically analogous to the summation wealth process used as an example in \autoref{sec:method}. More formally, we can denote the process $\psi_{t} = \psi_{t-1} + \gamma \, \left(\rz_{t-1} - \epsilon\right)$, and under the null (\autoref{eq:hypotheses-formulation}) it follows that $\mathbb{E}\left[\psi_{t} \ \vert \ \mathcal{F}_{t-1}\right] \leq \psi_{t-1}$ indicates a risk-controlling parameter, whereas under the alternative $\mathbb{E}[\rz_t | \mathcal{F}_{t-1}] > \epsilon$, and $\psi_t$ thus grows as a martingale accumulating evidence against the null. The `testing by betting' interpretation helps clarify the key distinction to our approach---how the designed wealth process is subsequently utilized. Whereas we take a testing decision on the basis of a rejection threshold, RRC does not enforce such a stopping rule but re-invests the wealth in an update step to dynamically adjust prediction set sizes. While offering a convenient step towards model adaptation, the rule is tied to the explicit monotonicity assumption underlying RRC, wherein a larger $\psi_t$ reduces the risk by enlargening the prediction set and vice versa. Such monotonic behaviour in the thresholds is desirable, but not always available.

\paragraph{Risk control under the i.i.d. data stream setting.} \looseness=-1In the simple case where the test stream originates \emph{i.i.d} $(\vx_t, \vy_t)_{t \in \gT} \sim P_0$ rather than from time-dependent distributions $P_t$, we obtain that $\gR_t(\psi) = \gR_0(\psi)$ is a time-\emph{independent} risk, and the independence between samples further simplifies the risk definition (we detail our argument in \autoref{app:math-iid-stream}). We may then conveniently reverse the hypotheses pair from \autoref{eq:hypotheses-formulation} to form the test  
\begin{align*}
    H_{0}(\psi): \exists t \in \gT: \gR_0(\psi) > \epsilon, \; H_{1}(\psi) : \gR_0(\psi) \leq \epsilon \; \forall t \in \gT
\end{align*}
and define a reverse wealth process as $M_{t}\left(\psi\right) = \prod_{i=1}^{t}\left(1 + \lambda_{i}\left(\epsilon - \rz_i\right)\right)$. The corresponding $\psi$-CS construction is given by $C_{t}^{\psi} = \{\psi \in \gPsi \ : \ M_{t}\left(\psi\right) \geq 1/\delta\}$. Crucially, we can exploit the fact that $P_0$ is static and thus any drawn test conclusions on \emph{non-violation} of the risk (now formalized in $H_1(\psi)$) hold indefinitely in the future. In other words, $C_{t}^{\psi}$ will only grow as more thresholds are found to be safe, but never shrink. This permits leveraging the Type-I error control property of the wealth process to claim strong \emph{time-uniform} risk control guarantees of the form $\mathbb{P}\left(\forall t \in \gT \ : \ \gR_{0}\left(\psi\right) \leq \epsilon\right) \geq 1 - \delta$, moving beyond static risk control assurances. This approach has been leveraged directly by \cite{xu2024active} to extend \emph{RCPS} \citep{bates2021distribution} to the stream setting, and indirectly by \cite{adaptiveltt} to make \emph{Learn-then-Test} \citep{angelopoulos2021learn} more adaptive. Naturally, such forward-looking assurances continue to only hold for the setting of a \emph{static} distribution $P_0$, and are not applicable in our challenging shift setting.

\section{Theoretical Analysis}
\label{sec:theory}

\looseness=-1Next, we establish the theoretical guarantees our approach enjoys. To summarize, we suggest monitoring the risk using the wealth process $M_t(\psi)$ (\autoref{eq:test-supermartingale}) to detect risk violations and raise a signal when the stopping rule $1/\delta$ is met. Running this procedure simultaneously for all candidates $\psi \in \gPsi$, a set of thresholds $C_t^{\psi}$ (\autoref{eq:psi-cs}) deemed non-violating is returned at every time step $t$. In turn, $\gPsi \backslash C_t^\psi$ denotes the thresholds for which a test decision on risk violation at time $t$ has been made. In the following series of statements we provide insights on the \emph{statistical validity} and \emph{efficiency} of our approach, deferring all proofs to \autoref{app:math-proofs}. We begin by stating the essential martingale properties of the process $M_{t}\left(\psi\right)$ which render it practical for risk monitoring.

\begin{lemma}[Valid wealth process]
\label{thm:valid-wealth-process}
    For any $\psi \in \gPsi$ such that $\mathbb{E}_{P_t} [\rz_t \mid \mathcal{F}_{t-1}] \leq \epsilon \; \forall t \in \mathcal{T}$ satisfies the null, the process $M_t(\psi)$ in \autoref{eq:test-supermartingale} is a valid test supermartingale for the predictable betting rate  $\lambda_t \in [0, 1/\epsilon)$.
\end{lemma}

\looseness=-1 This follows directly from satisfying the conditions of a valid test supermartingale (\autoref{app:math-defn.-and-terminology}), and $\lambda_t \in [0, 1/\epsilon)$ ensures that $M_{t}(\psi)$ remains non-negative with its expected value upper-bounded by $M_0 = 1$ under the considered null hypothesis $H_{0}\left(\psi\right)$ (\autoref{eq:hypotheses-formulation}). We next characterize the \emph{false alarm guarantee} that our procedure natively derives from \hyperref[thm:valid-wealth-process]{Lemma~\ref{thm:valid-wealth-process}} via its Type-I error control property, given as

\begin{lemma}[False alarm guarantee]
\label{thm:false-alarm}
    For any $\psi \in \gPsi$ such that  $\mathbb{E}_{P_t} [\rz_t \mid \mathcal{F}_{t-1}] \leq \epsilon \; \forall t \in \mathcal{T}$ satisfies the null, it holds that $\mathbb{P} \left( \exists t \in \mathcal{T} : M_t (\psi) \geq 1/\delta \right) \leq \delta.$
\end{lemma}

\looseness=-1The result ensures that a false positive or false alarm---when $M_t(\psi)$ crosses $1/\delta$ and \emph{incorrectly} signals risk violation as the threshold $\psi$ is in fact risk-controlling---occurs with at most probability $\delta$. Consequently, any threshold that does \emph{not} violate the risk level $\epsilon$ at time $t$ is guaranteed to be included in $C_{t}^{\psi}$ with high probability, \ie, it holds that ${\mathbb{P}_{H_0}(\forall t \in \mathcal{T} \ : \ M_{t}(\psi) < 1/\delta) > 1-\delta}$. Such high-probability safety assurances can be given under the null hypothesis $H_0(\psi)$ but do not translate to the alternative (see also \autoref{subsec:connection-methods} for the reverse hypotheses). Leveraging the properties of $M_t(\psi)$, what statements can be made with respect to $H_1(\psi)$, indicating true risk violation? First, we establish the method's asymptotic consistency under regular conditions.

\begin{lemma}[Asymptotic consistency]
\label{thm:power-one-property} 
    For any $\psi \in \gPsi$ such that $\mathbb{E}_{P_t}\left[\rz_t \ \vert \ \mathcal{F}_{t-1}\right] \leq \epsilon$ for finitely many steps $t \in \gT$ and $\mathbb{E}_{P_t}\left[\rz_t \ \vert \ \mathcal{F}_{t-1}\right] > \epsilon$ otherwise, it holds that ${\mathbb{P}(\tau(\psi) < \infty) = 1}$, where $\tau(\psi)$ denotes the stopping time.
\end{lemma}

\looseness=-1In words, our hypothesis test aligns with the classical \emph{sequential test of power one} \citep{darling1968some} and assures that a risk-violating threshold will be inevitably detected (with finite stopping time). While this ensures asymptotic correctness, a practical application demands more than \emph{eventual} detection---it requires efficiency in the evidence accumulation. We borrow such a notion of statistical efficiency by directly leveraging the property of \emph{growth rate optimality} to guide our betting rate $\lambda_t$. Informally, an adaptive strategy for the bets $(\lambda_t)_{t \in \mathcal{T}}$ can be designed to directly maximize the growth of the wealth process under the alternative, ensuring efficient evidence accumulation \citep{waudby2024estimating, grunwald2024safe}. We formally define the condition as

\begin{definition}[Growth rate optimality (GRO)] 
\label{thm:gro}
    The betting rate $\lambda_t$ is growth rate optimal if it satisfies the condition ${\lambda_t = \argmax_{\lambda \in [0, 1/\epsilon)} \mathbb{E}_{H_1}[\log M_t(\psi)]}$.
\end{definition}

\looseness=-1We follow the GRO principle in our experiments (see \autoref{sec:exp}), and hence ensure our procedure retains maximal efficiency (or statistical power) among all possible wealth processes. Finally, under some additional conditions we propose characterizing the method's stopping time behaviour, and by extension its \emph{detection delay} denoting the practical utility. Defining $\tau_{*}(\psi) = \inf\{t \in \gT \ : \ \mathbb{E}_{P_{t}}\left[\rz_t \ \vert \ \mathcal{F}_{t-1}\right] > \epsilon\}$ as the (unknown) \emph{true} stopping time of $\psi$ (\ie, when the true risk is violated), and $\tau(\psi) = \inf\{t \in \gT \ : \ M_{t}\left(\psi\right) \geq 1/\delta \}$ as our stopping signal, the method's detection delay is given by $\tau\left(\psi\right) - \tau_*\left(\psi\right)$. We propose the following characterization:

\begin{proposition}[Detection delay bound]
\label{thm:detection-delay-argument}
    A worst-case detection delay for the hypothesis pair in \autoref{eq:hypotheses-formulation} and wealth process $M_t(\psi)$ in \autoref{eq:test-supermartingale} is characterized by ${(\tau(\psi) - \tau_*(\psi)) \approx \gO((\log(1/\delta) \, + \, T)/(\lambda \cdot \mu))}$, where $\mu$ denotes the risk violation intensity and $T$ a shift changepoint.
    \vspace{-2mm}
\end{proposition}

% \vspace{-3mm}
\looseness=-1We elaborate on \hyperref[thm:detection-delay-argument]{Prop.~\ref{thm:detection-delay-argument}} in \autoref{app:math-proofs}, but can intuitively observe \emph{(i)} inverse proportionality to both the betting rate $\lambda$ and the violation strength $\mu$, indicating a faster detection when evidence grows adaptively and is strong; and \emph{(ii)} direct proportionality to $T$, indicating a slower detection when shifts occur later in the stream as the initial evidence favouring $H_0(\psi)$ needs to be overcome (slowing the wealth's growth).

\section{Related Work}
\label{sec:related-work}

\looseness=-1\cite{waudby2024estimating} offer an in-depth study on detecting deviations in the means of bounded quantities for stream settings, providing useful tools (\eg~in terms of betting rate design) for the testing of risks following \autoref{eq:def-risk}. Very recently, \cite{fan2025testing} explore some settings for additional bounds on the variance. Whereas they consider data to originate from a fixed source distribution $P$ with unknown mean $\mu$ to be estimated, our observations originate from variable, time-dependent distributions $P_t$, and we simultaneously monitor multiple means (corresponding to the threshold-dependent risks $\gR_t(\psi)$). Thus our underlying hypothesis dictating the test design is subtly, but distinctly different. Yet, strong results on the universal representation of test martingales (\eg~stated by \cite{waudby2024estimating}, Prop. 3) render the process structure in \autoref{eq:test-supermartingale} useful for a wide range of testing problems. We attempt to intuitively motivate this via a `forecasting game' in \autoref{sec:method}. 

\paragraph{Sequential testing and risk monitoring.} \looseness=-1 \cite{xu2024active, adaptiveltt} leverage above results on deviations in means to provide strong time-uniform risk control for \emph{i.i.d} streams, discussed further in \autoref{subsec:connection-methods}. Closely related to our work, \cite{podkopaev2021tracking} monitor a \emph{running risk} of the form $\gR_r(\psi) = \frac{1}{t}\sum_{i=1}^{t} \mathbb{E}_{P_i}[\rz_i]$ under the sequential testing framework, with similar false alarm guarantees. However, we consider the more challenging instantaneous true risk $\gR_t(\psi)$ at any given time step, which can recover $\gR_r(\psi)$ but not vice versa. Furthermore, their experimental design tends to distinguish between benign and harmful shifts caused by a dominant shift initiated at $P_0$ (akin to changepoint detection), whereas we incorporate a broader variety of shifts. Finally, we do not impose sample independence. Their approach was reformulated by \cite{amoukou2024sequential} for unlabelled streams, and relatedly \cite{bar2024protected} suggest an unsupervised covariate shift detector on the basis of entropy-matching. Particular to out-of-distribution detection, \cite{vishwakarma2024taming, sun2024online} also leverage martingale-based constructions.

\paragraph{Other sequential testing under shift.} \looseness=-1 Stream data denotes a particular test setting under the `testing-by-betting' framework, lending itself naturally to the use of test martingales\footnote{Also referred to therein as \emph{sequential anytime-valid inference}.} \citep{ramdas2023game, ramdas2024hypothesis}. Within that framework, previous work has considered a variety of testing problems, such as on exchangeability \citep{vovk2021testing, saha2024testing}, independence \citep{podkopaev2023sequential}, two-sample testing \citep{Shekhar2021NonparametricTT, PandevaBNF24, PandevaFRS24, luo2024online}, or changepoint detection \citep{shekhar2023sequential, shekhar2024reducing, vovk2021retrain, volkhonskiy2017inductive, shin2023detectors}. Theses works differ from ours in terms of the hypotheses they address, their data settings (\eg~by simultaneously observing two separate data streams), or experimental designs (\eg~detecting a single changepoint or shift). Additional related works on static risk control and extensions to stream settings not using sequential testing can be found in \autoref{app:background}.

\section{Empirical Results}
\label{sec:exp}

\begin{figure*}[t]
    \centering
    \includegraphics[width=1\linewidth]{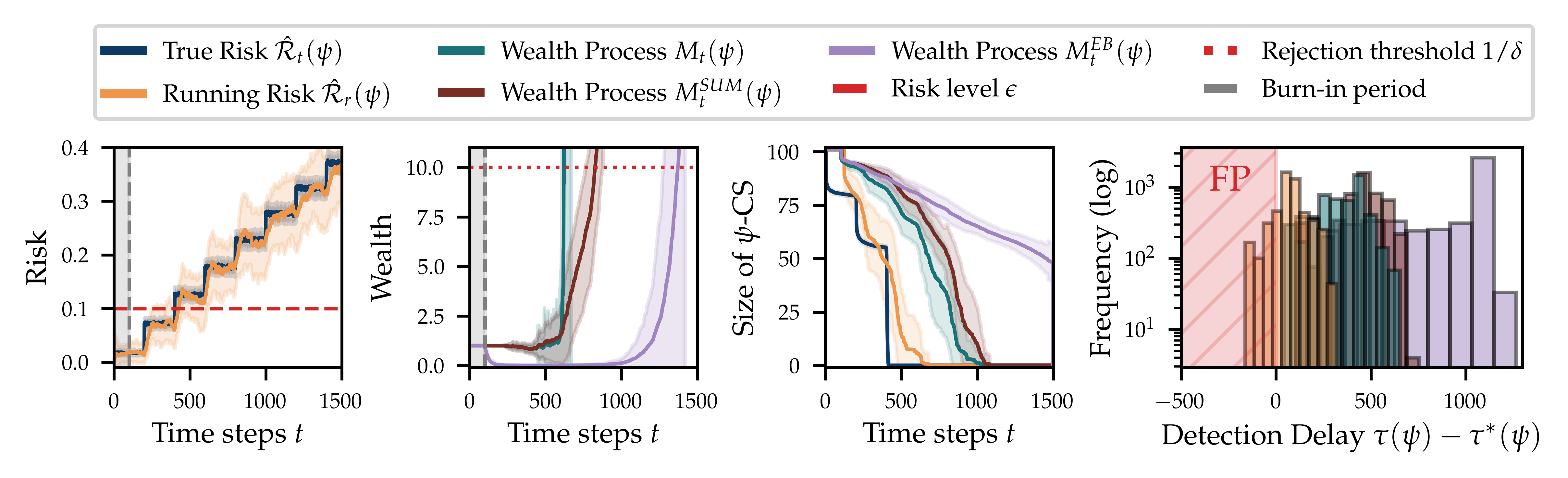}
    \caption{
    Results for \textbf{outlier detection with a stepwise shift} (\autoref{subsec:exp-ood}). \emph{From left to right:} Visuals of the growing risk and wealth process behaviour with respective rejection thresholds $\epsilon$ and $1/\delta$, for a single threshold candidate (here $\psi=0.50$); the behaviour of the valid threshold set $\psi$-CS (\autoref{eq:psi-cs}), which eventually shrinks to zero signalling a model update; and the empirical distributions of detection delays $\tau(\psi) - \tau_*(\psi)$ across all $\psi \in \gPsi$, including the false alarm region (FP). We also have $B=1, S=50$ and $t_{out} = 200$, with results evaluated over $R=50$ trials (mean and std. deviation).
    }
    \vspace{-3mm}
\label{fig:ood-exp}
\end{figure*}

\looseness=-1We empirically validate our risk monitoring approach on two tasks, outlier detection (\autoref{subsec:exp-ood}) and set prediction (\autoref{subsec:exp-sets}). The first experiment induces shifts by mixture sampling in order to demonstrate monitoring behaviour for explicit scenarios, while the latter is based on naturally occuring temporal shifts. Evaluating diverse, real-world datasets, we find that the method ensures both timely detection of risk violations and a controlled false alarm rate. We next outline our baselines and practical design choices, followed by each experiment in more detail (see also \autoref{app:sec-exp-design}). Our code is publicly available at \url{https://github.com/alextimans/risk-monitor}.

\paragraph{Baselines.} \looseness=-1We compare our primary monitoring approach, the wealth process $M_t(\psi)$ described by \autoref{eq:test-supermartingale}, to the following (empirical) risk tracking mechanisms:
\begin{itemize}[leftmargin=13pt, itemsep=0pt, topsep=0pt]
    \item[\emph{(i)}] An empirical estimate of the \emph{unobservable oracle} or true population risk $\gR_t(\psi)$, computed as ${\hat{\gR}_t(\psi) = \frac{1}{B_*} \sum_{b=1}^{B_*} z_{t, b}, \; z_{t, b} \sim P_t}$ for a batch draw $B_*$ of large size (\eg~$B_* = 1000$). We desire for $M_t(\psi)$ to emulate the monitoring behaviour of $\hat{\gR}_t(\psi)$ as closely as possible while controlling false alarms by \hyperref[thm:false-alarm]{Lemma~\ref{thm:false-alarm}}.
    
    \item[\emph{(ii)}] An empirical estimate of the running risk $\gR_r(\psi)$, accumulated over the data stream for a given time step $t$ as ${\hat{\gR}_r(\psi) = \frac{1}{t} \sum_{i=1}^{t} z_i}$. This is the risk quantity evaluated both by \cite{podkopaev2021tracking} as a tractable estimate of the running risk, and \cite{Feldman2022AchievingRC} directly as a rolling risk target (\autoref{subsec:connection-methods}). Note that since we are monitoring the more challenging instantaneous risk $\gR_t(\psi)$, the estimator $\hat{\gR}_r(\psi)$ is nominally void of any false alarm guarantees.
    
    \item[\emph{(iii)}] The summation wealth process illustrated in \autoref{sec:method}, given by ${M^{SUM}_{t}(\psi) = \sum_{i=1}^{t} \lambda_i (z_i - \epsilon)}$. This process retains the same false alarm guarantees as $M_t(\psi)$, but tends to be less adaptive as evidence is accumulated additively.
    
    \item[\emph{(iv)}] The \emph{predictably-mixed Empirical-Bernstein} wealth process from \cite{waudby2024estimating}, given by ${M^{EB}_{t}(\psi) = \prod_{i=1}^{t} \exp\{ \lambda_i \, (z_i - \epsilon) - v_i \, \rho(\lambda_i) \}}$ where $v_i = 4\,(z_i - \hat{\mu}_{i-1})^2$, $\rho(\lambda_i) = 1/4\,(- \log(1 - \lambda_i) - \lambda_i)$, and we use the \emph{predictable plug-in} betting rate $\lambda^{EB}_i = \min\left\{\sqrt{\frac{2 \, \log(2/\delta)}{\hat{\sigma}^2_{i-1} \, i \, \log(1 + i)}}, \frac{1}{2}\right\}$. A similar method is also derived by \cite{podkopaev2021tracking} to estimate confidence bounds on $\gR_r(\psi)$ in their problem setting, but we employ its direct form as a sequential test.
\end{itemize}

\paragraph{Choice of betting rate.} \looseness=-1We follow the growth rate optimality (GRO) condition outlined in \hyperref[thm:gro]{Definition~\ref{thm:gro}} to guide our choice of betting rate. Whereas selecting $\lambda_t$ based on direct wealth maximization is possible, the approach can be computationally expensive to re-evaluate for every candidate $\psi$ and time step $t$. Instead, we leverage a suggested approximation by \cite{waudby2024estimating}, yielding the closed-form expression 
\begin{equation*}
    \lambda^{AGR}_t = \max \left\{ 0, \min \left\{ \frac{\hat{\mu}_{t-1} - \epsilon}{\hat{\sigma}^2_{t-1} + (\hat{\mu}_{t-1} - \epsilon)^2}, \frac{1/2}{\epsilon} \right\} \right\},
\end{equation*}
where $\hat{\mu}_{t-1}$ and $\hat{\sigma}^2_{t-1}$ denote the estimated running mean and variance over $\{z_i\}_{i=1}^{t-1}$. Intuitively, the betting rate increases when the running mean is far from $\epsilon$, and is further amplified by a small variance. $\lambda^{AGR}_t$ is \emph{approximately} GRO \citep{shekhar2023near} and performs empirically similar to direct maximization. A range of other suitable bets is discussed in \cite{waudby2024estimating}, and we briefly touch upon this in \autoref{app:sec-exp-design}.

\paragraph{Batching, sliding window and burn-in.} \looseness=-1Instead of a data stream where samples arrive individually at every time step, we may also consider the arrival of small batches of size $B \ll B_*$, \ie~we sample $\{ (\vx_{t,b}, \vy_{t,b})\}_{b=1}^{B} \sim P_t$. The batch-wise evidence at every time step can be easily aggregated by, for instance, averaging, which tends to both reduce the variance of the tracking process and improve the detection delay $\tau(\psi) - \tau_*(\psi)$ with respect to the true risk even for small batches ($B=10$). Similarly, delays can be reduced by enhancing the adaptivity of any tracker via a sliding window of size $S$, wherein only the most recent observations for time steps $i \in [t-S, t]$ are considered. Intuitively, the observational history is truncated by discarding past information deemed irrelevant for the current shift environment. This renders the tracking process more reactive (\eg, via the betting rate parameters $\hat{\mu}_{t-1}, \hat{\sigma}^2_{t-1}$) but also increases sensitivity to the retained samples, heightening the chance of false alarms if the resulting evidence is misleading. The choice of $B$ and $S$ can sometimes be delicate, and results for different combinations are provided in \autoref{app:sec-exp-res}. Finally, we introduce an initial number of \emph{burn-in} time steps $t_{burn} = \left\lfloor 100/B \right\rfloor$ during which any risk tracker (aside of the true risk) does not test for risk violation but merely accumulates samples, in order to stabilize any running quantities such as $\hat{\gR}_r(\psi)$.

\begin{figure*}[t]
    \centering
    \includegraphics[width=1\linewidth]{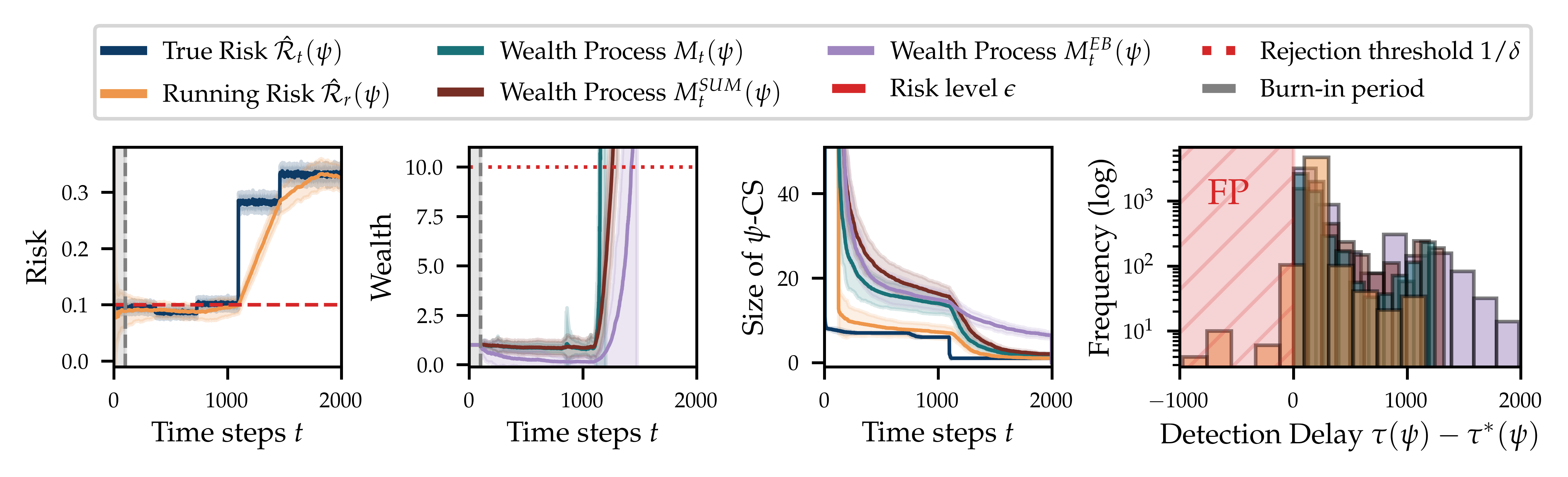}
    \caption{
    Results for \textbf{set prediction with a temporal shift on FMoW} (\autoref{subsec:exp-sets}). \emph{From left to right:} Visuals of the growing risk and wealth process behaviour with respective rejection thresholds $\epsilon$ and $1/\delta$, for a single threshold candidate (here $\psi=0.08$); the behaviour of the valid threshold set $\psi$-CS (\autoref{eq:psi-cs}), which eventually tends to zero signalling a model update; and the empirical distributions of detection delays $\tau(\psi) - \tau_*(\psi)$ across all $\psi \in \gPsi$, including the false alarm region (FP). We also have $B=1$ and $S=365$ (one year), with results evaluated over $R=50$ trials (mean and std. deviation).
    }
    \vspace{-3mm}
\label{fig:set-exp}
\end{figure*}

\subsection{Monitoring the Total Error Rate for Outlier Detection}
\label{subsec:exp-ood}

\looseness=-1We first consider the task of outlier detection, and instantiate the threshold predictor from \autoref{eq:def-threshold-pred} as ${\hat{f}_{\psi}(\rvx) = \mathbbm{1}[\texttt{out}(\rvx) \geq \psi]}$, where $\texttt{out}: \gS \rightarrow [0,1]$ maps the predictor's output to a bounded outlier score and $\mathbbm{1}[\cdot]$ is the indicator function. When $\texttt{out}(\rvx) \geq \psi$ evaluates true we declare the sample an outlier. Given a classification setting, we define $\texttt{out}$ as the normalized entropy of the base model's predictive distribution $\hat{p}(\rvy \mid \rvx)$\footnote{This is merely one possible choice, and can be easily swapped for other scoring mechanisms satisfying the required bounds.}. The target risk to monitor is given by the \emph{total error rate}, accounting for both cases of inlier (false positives, FP) and outlier misclassification (false negatives, FN) via the loss variable
\begin{equation*}
\label{eq:ood-ter}
    \rz_t = 
    \begin{cases}
        1, & \text{if } \texttt{out}(\rvx_t) \geq \psi \; \text{ and }\; (\rvx_t, \rvy_t) \sim P_{in}, \hfill \text{ (FP)} \\
        1, & \text{if } \texttt{out}(\rvx_t) < \psi \; \text{ and }\; (\rvx_t, \rvy_t) \sim P_{out}, \; \text{ (FN)} \\
        0 & \text{else.}
    \end{cases}
\end{equation*}
$P_{in}$ and $P_{out}$ denote inlier and outlier distributions, and the shifting stream is characterized by a time-dependent outlier probability $\pi_{t}^{out}$ such that $(\vx_t, \vy_t) \sim (1 - \pi_{t}^{out})\,P_{in} + \pi_{t}^{out}\,P_{out}$ for $t \in \gT$ is generated by mixture sampling. We consider three distinct shift settings: \emph{(i)} an \emph{i.i.d} stream where trivially $\pi_{t}^{out} = 0$ across all time steps; \emph{(ii)} an immediate stark outlier shift where $\pi_{t}^{out} = 1$ early on; and \emph{(iii)} a stepwise shift with $\pi_{t}^{out} \in \{0, 0.05, 0.1, \dots, 1 \}$ increasing every $t_{out}$ time steps. Risk parameters are set to common values $\epsilon = 0.1, \delta = 0.1$, and we simulate for $T = 1500$ steps. $P_{in}$ and $P_{out}$ are given by CIFAR-10 \citep{krizhevsky2009learning} and SVHN \citep{netzer2011reading} respectively, with a base classifier (ResNet-50) trained on CIFAR-10.

\looseness=-1Our results in \autoref{fig:ood-exp} for the stepwise shift assert that as the shift intensity increases, so does the number of risk-violating thresholds, leading to a gradual shrinkage of the $\psi$-CS towards zero. Among risk trackers the running risk $\hat{\gR}_r(\psi)$ emulates the true risk well but tends to misinterpret evidence, resulting in an undesirable number of false alarms. In contrast, all martingale-based trackers uphold the guarantee, at the cost of increased detection delays. Among them, the monitoring behaviour of the wealth process $M_t(\psi)$ most closely aligns with the true risk, striking a good trade-off. Results for other shift settings can be found in \autoref{app:sec-exp-res}, where as anticipated \emph{(i)} for the \emph{i.i.d} case most thresholds remain valid and the $\psi$-CS stabilizes over the full data stream; and \emph{(ii)} for the immediate shift all thresholds are rejected as soon as possible, correctly identifying $\hat{f}_{\psi}$ as highly unreliable.

\subsection{Monitoring the Miscoverage Rate for Set Prediction}
\label{subsec:exp-sets}

\looseness=-1Next we consider set prediction tasks on data subject to \emph{natural temporal shifts}, both for the classification and regression setting.

\begin{figure*}[t]
    \centering
    \includegraphics[width=1\linewidth]{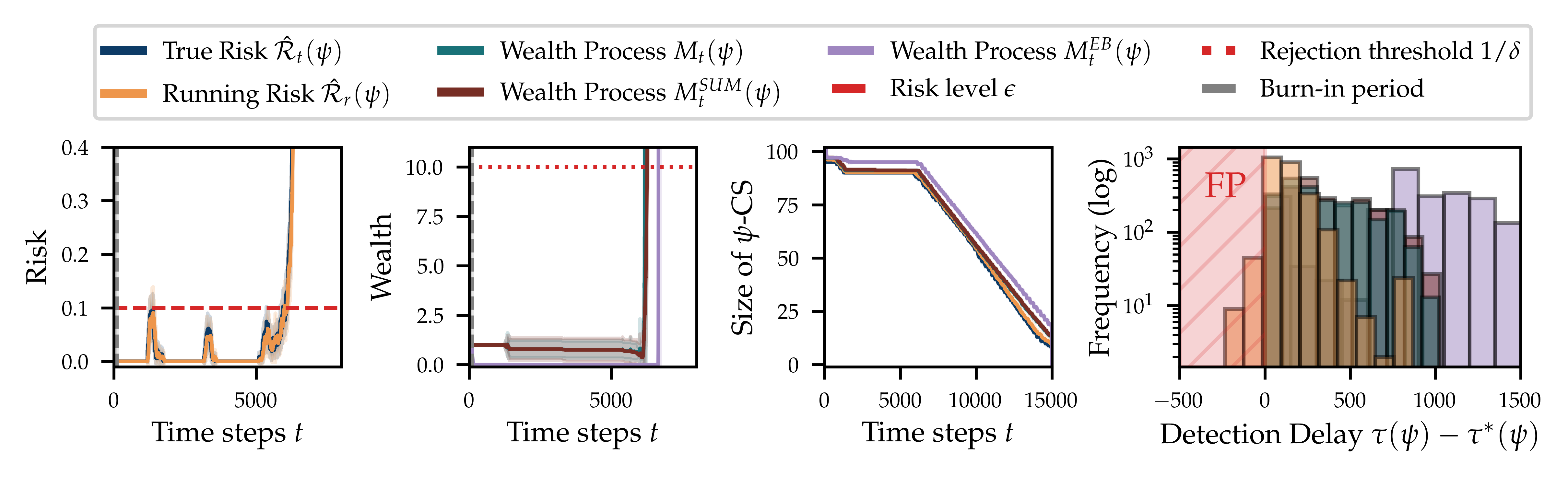}
    \caption{
    Results for \textbf{set prediction with a temporal shift on Naval propulsion} (\autoref{subsec:exp-sets}). \emph{From left to right:} Visuals of the growing risk and wealth process behaviour with respective rejection thresholds $\epsilon$ and $1/\delta$, for a single threshold candidate (here $\psi=0.005$); the behaviour of the valid threshold set $\psi$-CS (\autoref{eq:psi-cs}), which eventually tends to zero signalling a model update; and the empirical distributions of detection delays $\tau(\psi) - \tau_*(\psi)$ across all $\psi \in \gH$, including the false alarm region (FP). We also have $B=1$ and $S=50$, with results evaluated over $R=50$ trials (mean and std. deviation).
    }
    \vspace{-3mm}
\label{fig:set-exp-naval}
\end{figure*}

\paragraph{Functional Map of the World.}\looseness=-1For classification, we instantiate $\hat{f}_{\psi}$ as a set predictor of the form 
\begin{equation*}
\label{eq:set-pred}
    \hat{f}_{\psi}(\rvx) = \{ \vy \in \gY: \hat{p}(\rvy = \vy \mid \rvx) \geq \psi \},
\end{equation*}
and the base classifer once more returns a predictive distribution $\hat{p}(\rvy \mid \rvx)$ used to determine class inclusion in the set. A natural risk to monitor here is the \emph{miscoverage rate} with loss variable $\rz_t = \mathbbm{1}[\vy_t \notin \hat{f}_{\psi}(\rvx_t)]$, where $\vy_t$ denotes the true label. We consider the \emph{Functional Map of the World} dataset (FMoW) \citep{christie2018functional}, a large-scale satellite image dataset on building and land use over 16 years, and employ a time-dependent partitioning proposed by \cite{yao2022wild}. Therein a natural shift occurs as the \emph{same} satellite image locations capture land use changes over time. The classifier (DenseNet-121) is trained on the first 11 years, and we increase the test stream frequency by sampling chronologically from the final five years every 365 time steps (simulating daily observations). We again set $\epsilon = 0.1, \delta = 0.1$ and run for $T = 2000$ steps. 

\looseness=-1 Our results in \autoref{fig:set-exp} draw similar conclusions as in \autoref{subsec:exp-ood}, that is, the proposed wealth process $M_t(\psi)$ produces the lowest detection delays among all risk monitoring processes with false alarm control, while the running risk prematurely rejects some threshold candidates. Interestingly, the natural temporal shift induces a non-monotonic risk profile, wherein miscoverage for the second year slightly drops, but thereafter starkly increases. We elaborate on the connection between risk profiles and threshold behaviour in \autoref{app:sec-exp-design}, and provide complete results in \autoref{tab:app-fmow-natural}.

\paragraph{Naval propulsion system.}\looseness=-1For regression, the set predictor takes the interval form ${\hat{f}_{\psi}(\rvx) = [\hat{f}(\rvx) - \psi, \, \hat{f}(\rvx) + \psi]}$, with $\hat{f}(\rvx)$ returning point estimates. The target risk remains the miscoverage rate, and we consider predictive maintenance data on naval gas turbine behaviour \citep{cipollini2018condition}. This tabular time series consists of $\sim 12 \, 000$ recordings for various turbine system parameters, and an associated turbine compressor degradation coefficient denoting the compressor’s health. Over time this degradation coefficient steadily increases, denoting a gradual equipment decay. We train a Random Forest regressor on the initial `healthy' compressor state (enriched via jittered resampling) which expectedly fails to extrapolate as the degradation worsens, resulting in decreased performance in line with a temporal shift. Once more we have $\epsilon = 0.1, \delta = 0.1$ and run our monitoring process for the full time series.

\looseness=-1Our results in \autoref{fig:set-exp-naval} and summarized in \autoref{tab:app-uci-natural} draw consistent conclusions with other experiments. Specifically, we observe \emph{(i)} incurred false positives by the running risk, in particular for more adaptive tracking windows (smaller $S$); and \emph{(ii)} lowest detection delays for the wealth process $M_t(\psi)$ among all trackers with false alarm guarantees. Furthermore, the gap between running risk and wealth process remains fairy narrow under most realistic settings (\ie, small $B$ and large $S$). Visually, the $\psi$-CS (\autoref{eq:psi-cs}) stabilizes during the initial healthy state, but consistently shrinks as turbine compressor degradation and thus distributional shift worsens. The visualized threshold, being very small, displays sensitive risk behaviour even during early time steps.

\vspace{-3mm}
\section{Discussion}
\label{sec:discussion}

\looseness=-1We investigate sequential testing-based approaches to monitor an unobservable, time-dependent bounded risk $\gR_t(\psi)$ in a dynamic data stream setting, challenged by unknown and repeated distribution shifts. Motivated by the `testing by betting' framework \citep{waudby2024estimating}, our martingale-based monitoring process ensures timely detection of risk violations whilst providing finite-sample control over the false alarm rate. This renders a statistically rigorous and yet practical procedure for risk monitoring.

\looseness=-1However, we are inherently limited in our safety assurances by the unpredictability of any occuring shift, and the minimal assumptions we impose on it. More informative forward-looking assurances can potentially be obtained if additional restrictions are considered, such as constraints on the shift origin or its intensity and growth rate. Similarly, rephrasing our problem statement in terms of a different hypothesis might simplify the task and offer more efficient or \emph{unsupervised} monitoring, \eg~by drawing inspiration from \cite{bar2024protected}'s entropy-matching idea or \cite{amoukou2024sequential}'s label-free quantile test. Possible unsupervised extensions may include recasting the task as two-sample testing \citep{PandevaBNF24,PandevaFRS24}, using generalization estimation \citep{baek2022agreement, rosenfeld2023almost}, or leveraging calibration properties \citep{gupta2020distribution}. Similarly, the model update step can be integrated into the framework, \eg~via test-time adapation \citep{ttasurvey} or online and continual learning principles \citep{wang2024comprehensive}. Ultimately, the provision of practical safety assurances for robust model behaviour at deployment \emph{under arbitrary shift} is a challenging problem \citep{fang2022out}.

% Acknowledgments, Societal Impact

\begin{contributions} 
    AT and RV contributed together to ideation and methodology. AT developed and conducted all experiments and led paper writing, while RV initiated the problem statement and the preliminary approach, developed the theoretical motivation and proofs and co-authored sections of the paper. EN provided project guidance and general feedback. CN contributed to ideation, theoretical development, project guidance and general feedback.
\end{contributions}

\begin{acknowledgements}
    We thank Teodora Pandeva for initial discussions that helped shape the project. Additionally, we thank Wouter Koolen and Yaniv Romano for insightful discussions. This project was generously supported by the Bosch Center for Artificial Intelligence. Eric Nalisnick did not use any resources from Johns Hopkins for this work.
\end{acknowledgements}

% References
\bibliography{main}

%%%%%%%%%%%%%%%%%%%%%%%%%%%%%%%%
% APPENDIX
%%%%%%%%%%%%%%%%%%%%%%%%%%%%%%%%

\clearpage
\appendix
\onecolumn
\begin{center}
    {\Large\bfseries On Continuous Monitoring of Risk Violations under Unknown Shift}\\[0.5em]
    {\Large\bfseries --- Supplementary Material ---}
\end{center}
\label{appendix}
\tableofcontents
\newpage

\section{Mathematical Details}
\label{app:math}

We provide relevant mathematical details to complement the main text, including \emph{(i)} on the terminology of (super)martingales and the associated measure-theoretic objects, \emph{(ii)} a more detailed description of the summation wealth process, \emph{(iii)} insights on risk control under the \emph{i.i.d} data stream setting, and finally \emph{(iv)} formal proofs for our main theoretical statements.

\subsection{Definitions and Terminology}
\label{app:math-defn.-and-terminology}

Given a sequence of random variables $\rvu^{t} = (\rvu_1, \rvu_2, \ldots, \rvu_t)$ we denote the smallest $\sigma$-field generated by $\rvu^{t}$ by $\mathcal{F}_{t} = \sigma\left(\rvu^{t}\right)$. An indefinitely running sequence then leads to the filtration $\mathcal{F} = (\mathcal{F}_{t})_{t=0}^{\infty}$, defined as the increasing sequence of generated $\sigma$-fields $\mathcal{F}_{0} \subset \mathcal{F}_{1} \subset \mathcal{F}_{2} \subset \cdots$, where $\mathcal{F}_{0}$ is the trivial $\sigma$-field. A sequence of random variables $(M_t)_{t=0}^{\infty}$ is called a \emph{martingale} if it is adapted to the filtration $\mathcal{F}$, \ie~each $M_t$ is $\mathcal{F}_{t}$-measurable, integrable, and satisfies $\mathbb{E}\left[M_{t} \ \vert  \ \mathcal{F}_{t-1}\right] = M_{t-1}$. If the equality is replaced by an inequality ($\leq$) then we call $(M_t)_{t=0}^{\infty}$ a \emph{supermartingale}. Furthermore, we define a sequence $(\lambda_{t})_{t=0}^{\infty}$ to be \emph{predictable} if $\lambda_{t}$ is $\mathcal{F}_{t-1}$-measurable, meaning $\lambda_{t}$ may only depend on past information obtainable up to and including time $t-1$. We also define the random variable  $\tau: \Omega \to \mathbb{N} \cup \{\infty\}$ as the \emph{stopping time} with respect to filtration $\mathcal{F}$ if, for every step $t \geq 0$, the event $\{\tau \leq t\}$ is included in the $\sigma$-field $\mathcal{F}_t$, \ie~$\{\tau \leq t\} \in \mathcal{F}_t$. This ensures the stopping decision at time $t$ to be based solely on previous information, meaning $\tau$ also cannot `peek into the future'. 

Finally, we introduce the below martingale concentration inequality subsequently employed in some of our proofs.
\begin{lemma}[Azuma-Hoeffding Inequality \citep{Hoeffding1994}]
\label{thm:azuma-hoeffding}
      Let $(\rv_i)_{i=1}^{t}$ be a martingale difference sequence adapted to the filtration $(\mathcal{F}_i)_{i=0}^{t}$, meaning that $\mathbb{E}[\rv_i \,|\, \mathcal{F}_{i-1}] = 0 \; \forall i$. Suppose there exist constants $c_i$ such that for all $i = 1, \dots, t$ we have $|\rv_i| \leq c_i$ almost surely. Then for any $\eta > 0$ it holds that \[\mathbb{P} \left( \left| \sum_{i=1}^{t} \rv_i \right| \geq \eta \right) \leq 2\exp\left(-\frac{\eta^2}{2 \sum_{i=1}^{t} c_i^2} \right),\] and similarly in the one-sided cases it holds that \[\mathbb{P}\left(\sum_{i=1}^{t} \rv_i \geq \eta \right) \leq \exp \left( -\frac{\eta^2}{2 \sum_{i=1}^{t} c_i^2} \right) \quad \text{as well as} \quad \mathbb{P} \left( \sum_{i=1}^{t} \rv_i \leq -\eta \right) \leq \exp \left( -\frac{\eta^2}{2 \sum_{i=1}^{t} c_i^2} \right).\]
\end{lemma}

\subsection{The Summation Wealth Process}
\label{app:sum-process}

As motivated in \autoref{sec:method}, the summation wealth process is given by $M_{t}\left(\psi\right) = \sum_{i=1}^{t}\lambda_i \left(\rz_i - \epsilon\right)$, with $(\lambda_t)_{t \in \gT}$ being the predictable betting rate. It is straightforward to see that under the null $H_{0}\left(\psi\right)$ (\autoref{eq:hypotheses-formulation}) this sequence forms a supermartingale, and hence does not grow meaningfully. Conversely, evidence for risk violation is collected \emph{iff} $M_{t}\left(\psi\right)$ is not a supermartingale. For the edge case where $\mathbb{E}_{P_t}\left[\rz_{t} \ \vert \ \mathcal{F}_{t-1}\right] = \epsilon$ and $M_{t}\left(\psi\right)$ only forms a martingale sequence we may apply the one-sided case of Lemma~\ref{thm:azuma-hoeffding} to bound the martingale's growth beyond some value $\eta$ with high probability. Given such a loss sequence $(\tilde{\rz}_{t})_{t \in \gT}$ `on the edge' of risk violation, we can argue that $\sum_{i=1}^{t}\lambda_i\left(\tilde{\rz}_{t} - \epsilon\right) \geq \sum_{i=1}^{t}\lambda_i\left(\rz_{t} - \epsilon\right)$ and hence the bound extends to $M_{t}\left(\psi\right)$ as 
\[\mathbb{P}\left(\sum_{i=1}^{t}\lambda_i \left(\rz_i - \epsilon\right) \geq \eta\right) \leq \exp\left(-\frac{\eta^{2}}{2t}\right),\] where the boundedness of $\rz \in \gL, \; \gL \subseteq [0,1]$ and $\epsilon \in (0,1)$ result in a bounded difference sequence and in particular $|\rz_t - \epsilon| \leq 1$. Choosing the particular limit value to be $\eta = \sqrt{2t \cdot \log (1/\delta)}$ we can see that $\mathbb{P}\left(M_{t}\left(\psi\right) \geq \eta\right) \leq \delta$, and thus growth beyond $\eta$ (and with rate $\gO(\sqrt{t})$) is contained with high probability $1-\delta$ for $\delta \in (0,1)$.

\subsection{Risk Control under the \emph{i.i.d.} Data Stream Setting}
\label{app:math-iid-stream}

Consider the simpler setting where the test stream originates \emph{i.i.d} from a non-shifting test distribution as $(\vx_t, \vy_t)_{t \in \gT} \sim P_0$. The risk quantity to monitor from \autoref{eq:hypotheses-formulation} directly simplifies due to independence as $\mathbb{E}_{P_t}\left[\rz_t \ \vert \ \mathcal{F}_{t-1}\right] = \mathbb{E}_{P_t}\left[\rz_t\right] = \gR_t(\psi)$, and since $P_0 = P_t \; \forall t \in \gT$ is static we have $\gR_t(\psi) = \gR_0(\psi)$ as a time-independent risk. We can now conveniently reverse our hypotheses under the sequential testing framework to exploit the fact that $P_0$ is static, and significant discoveries will thus hold even under future observations. That is, we can test for risk control directly by the hypothesis pair
\begin{align*}
\label{eq:hypotheses-iid}
    H_{0}(\psi): \exists t \in \gT: \gR_0(\psi) > \epsilon, \qquad H_{1}(\psi) : \gR_0(\psi) \leq \epsilon \; \forall t \in \gT,
\end{align*}
and use the following (reversed) wealth process and $\psi$-CS construction:
\begin{equation*}
\label{eq:eprocess-iid}
    M_t(\psi) = \prod_{i=1}^{t}\left(1 + \lambda_{i}\left(\epsilon - \rz_i \right)\right) \quad \text{ and } \quad C_t^{\psi} = \{ \psi \in \gPsi: M_t(\psi) \geq 1/\delta \}. 
\end{equation*}
Observe how once sufficient evidence is collected to support that the risk associated with a particular candidate $\psi$ does not exceed the tolerated risk level $\epsilon$, that candidate $\psi$ can be added to $C_t^{\psi}$ safely and indefinitely since the evidence collected remains meaningful under a static $P_0$. We can then directly leverage the Type-I error control property under the sequential testing framework \citep{ramdas2024hypothesis} (via Ville's Inequality) to state strong \emph{time-uniform} or \emph{anytime-valid} risk control guarantees. Specifically, it follows that for every $\psi \in \gPsi$ we have
\begin{equation*}
\label{eq:risk-control-iid}
        \mathbb{P}_{H_0}(\exists t \in \gT: M_t(\psi) \geq 1/\delta) \leq \delta \;
        \Rightarrow \; \mathbb{P}_{H_0}(\exists t \in \gT: \gR_0(\psi) \leq \epsilon) \leq \delta \;
        \Rightarrow \; \mathbb{P}(\forall t \in \gT: \gR_0(\psi) \leq \epsilon) \geq 1 - \delta.
\end{equation*}
In words, the probability of claiming risk control ($H_1(\psi)$) under risk violation ($H_0(\psi)$) is upper bounded by $\delta$, whereas under risk control we may perhaps mistakingly claim violation (and thus be overly conservative by excluding the associated $\psi$) but will not invalidate the risk level $\epsilon$. Thus, the overall probability of risk violation is controlled at level $1 - \delta$, rendering a strong safety assurance. Since $\gR_t(\psi)$ is not truly time-dependent neither is $C_t^{\psi}$, which will initially grow as evidence for each $\psi$ is collected and a decision on inclusion is made, and eventually stabilize. It is then straightforward to also recommend a particular threshold choice if the risk profile is monotonic or in some sense predictable, such as $\hat{\psi}_t = \min C_t^{\psi}$ as the least conservative threshold in case of a monotonically increasing risk. In other words, $\hat{\psi}_t$ will quickly tend to an optimal fixed choice $\hat{\psi}$ after a sufficient number of observations are processed, in contrast to our time-dependent shift setting.

\subsection{Proofs}
\label{app:math-proofs}

We provide proofs for the theoretical statements made in \autoref{sec:theory} below, restating each result for self-containment first.

\subsection*{Proof of Lemma~\ref{thm:valid-wealth-process}.}

\noindent\textbf{Lemma~\ref{thm:valid-wealth-process}.} (Valid wealth process). 
\emph{
    For any $\psi \in \gPsi$ such that $\mathbb{E}_{P_t} [\rz_t \mid \mathcal{F}_{t-1}] \leq \epsilon \; \forall t \in \mathcal{T}$ satisfies the null, the process $M_t(\psi)$ in \autoref{eq:test-supermartingale} is a valid test supermartingale for the predictable betting rate  $\lambda_t \in [0, 1/\epsilon)$.
}
\begin{proof}
    For any $\psi \in \gPsi$ the multiplicative wealth process takes the form $M_{t}\left(\psi\right) = \prod_{i=1}^{t}\left(1 + \lambda_{i}\left(\rz_i - \epsilon\right)\right)$, with $(\lambda_t)_{t \in \gT}$ the predictable betting rate. As we have the boundedness of $\rz \in \gL, \; \gL \subseteq [0,1]$ and $\epsilon \in (0,1)$ (from \autoref{sec:background}), restricting the betting rate to $\lambda_{t} \in [0, 1/\epsilon)$ ensures the term $1 + \lambda_t \left(\rz_t - \epsilon\right)$ to be non-negative. In addition, $M_{t}\left(\psi\right)$ is $\mathcal{F}_{t}$-measurable due to predictability of $\lambda_t$, and integrable due to the boundedness of $\rz$. As conditioned on $\mathcal{F}_{t-1}$ randomness originates in $\rz_t$, we also have that $\mathbb{E}_{P_t}\left[M_{t}\left(\psi\right) \,|\, \mathcal{F}_{t-1}\right] = M_{t-1}\left(\psi\right) + \lambda_{t}\cdot M_{t-1}\left(\psi\right)\cdot \mathbb{E}_{P_t}\left[(\rz_t - \epsilon) \, | \, \mathcal{F}_{t-1}\right] \leq M_{t-1}\left(\psi\right)$, satisfying the supermartingale condition. The final inequality follows from having $\mathbb{E}_{P_t}\left[(\rz_t - \epsilon) \,|\, \mathcal{F}_{t-1}\right] \leq 0$ for any risk-controlled threshold $\psi$. Hence, $M_{t}\left(\psi\right)$ forms a valid \emph{test supermartingale} or wealth process for $\psi$. Similarly, one can verify the condition in the reverse direction, \ie~that if $M_t (\psi)$ is valid, then it follows that $\mathbb{E}_{P_t} [\rz_t \, | \, \mathcal{F}_{t-1}] \leq \epsilon \; \forall t \in \mathcal{T}$ holds.
\end{proof}

\vspace{-5mm}
\paragraph{Remark.} Our general problem setting does not impose any restrictions on the nature of the data stream, \ie~arbitrary dependencies between time-dependent distributions $P_t$ and $P_{t'}, \; t \neq t'$ may persist. However, the inclusion of an \emph{independence} assumption on samples drawn from $P_{t}$ and $P_{t'}$ can simplify the considered hypothesis pair (\autoref{eq:hypotheses-formulation}), as now $\mathbb{E}_{P_t}\left[\rz_t \ \vert \ \mathcal{F}_{t-1}\right] = \mathbb{E}_{P_t}\left[\rz_t\right] = \gR_t(\psi)$ directly matches the time-dependent risk definition. For example, this assumption is leveraged in the \emph{i.i.d.} setting detailed above (\autoref{app:math-iid-stream}).

\subsection*{Proof of Lemma~\ref{thm:false-alarm}.}

\noindent\textbf{Lemma~\ref{thm:false-alarm}.} (False alarm guarantee). 
\emph{
    For any $\psi \in \gPsi$ such that  $\mathbb{E}_{P_t} [\rz_t \mid \mathcal{F}_{t-1}] \leq \epsilon \; \forall t \in \gT$ satisfies the null, it holds that $\mathbb{P} \left( \exists t \in \mathcal{T} : M_t (\psi) \geq 1/\delta \right) \leq \delta.$
}
\begin{proof}
    The above statement is a direct consequence of our test supermartingale's properties using Ville's inequality, which we state for completion.
    
    \noindent\textbf{Ville's inequality} \citep{ville1939etude}\textbf{.} Given a non-negative supermartingale sequence $(M_{t})_{t \in \gT}$ with $M_0 = 1$, it holds that $\mathbb{P} \left( \exists t \in \gT : M_t \geq 1/\delta \right) \leq \frac{\mathbb{E}[M_0]}{1/\delta} = \delta$.
    
    Similar to Lemma~\ref{thm:azuma-hoeffding} the inequality provides probabilistic assurances on the growth of the process, interpretable under the sequential testing framework \citep{ramdas2024hypothesis} as a Type-I error control property. Since Lemma~\ref{thm:valid-wealth-process} asserts that $M_{t}\left(\psi\right)$ is a valid supermartingale for any $\psi \in \gPsi$ for which $\mathbb{E}_{P_t} [\rz_t \mid \mathcal{F}_{t-1}] \leq \epsilon \; \forall t \in \mathcal{T}$ holds, our \emph{false alarm guarantee} is merely an interpretation of the obtained property in our problem setting.
\end{proof}

\subsection*{Proof of Lemma~\ref{thm:power-one-property}.}

\noindent\textbf{Lemma~\ref{thm:power-one-property}.}  (Asymptotic consistency).
\emph{
    For any $\psi \in \gPsi$ such that $\mathbb{E}_{P_t}\left[\rz_t \ \vert \ \mathcal{F}_{t-1}\right] \leq \epsilon$ for finitely many steps $t \in \gT$ and $\mathbb{E}_{P_t}\left[\rz_t \ \vert \ \mathcal{F}_{t-1}\right] > \epsilon$ otherwise, it holds that ${\mathbb{P}(\tau(\psi) < \infty) = 1}$, where $\tau(\psi)$ denotes the stopping time.
}
\begin{proof}
    The statement follows from the fact that our hypothesis test aligns with the properties of a classical \emph{sequential test of power one} \citep{darling1968some}, assuring eventual detection. However, we next provide an explicit proof based on the growth rate of $M_{t}\left(\psi\right)$ under the alternative ($H_{1}\left(\psi\right)$), closely following ideas outlined by \citet{PandevaFRS24} (in particular their consistency results, cf. Prop. 4.2). 
    
    With the stopping time defined in \autoref{app:math-defn.-and-terminology}, we first observe that $\mathbb{P}(\{\tau = \infty\}) = \mathbb{P}(\{\cap_{t\geq 1}\{\tau > t\}\}) \leq \mathbb{P}(\{\tau > t\})$. Taking the limit, we then also have $\mathbb{P}(\{\tau = \infty\}) \leq \limsup_{t \rightarrow \infty}\mathbb{P}(\{\tau > t\})$. Next, we argue why the limit ${\lim\sup_{t \rightarrow \infty}\mathbb{P}(\{\tau > t\})}$ tends to zero \emph{almost surely} under $H_1(\psi)$. For notational clarity we suppress all dependency on $\psi$ henceforth, thus denote the wealth process $M_t = \prod_{i=1}^{t}\left(1 + \lambda_i \cdot \delta_i \right)$, where $\delta_t = z_t - \epsilon$. We additionally introduce $S_t = \log M_t = \sum_{i=1}^{t}v_i$ as the logarithm of the wealth process, with cumulative terms $v_t = \log (1 + \lambda_t \cdot \delta_t)$. We also abbreviate the expectation of the term at some time $t$ as $A_t = \mathbb{E}[\rv_t \, | \, \gF_{t-1}]$. 

    \noindent\emph{The stopping condition.} With all this notation in place, let us consider the event probability $\mathbb{P}(\{\tau > t\})$, \ie~the probability of observing a stopping time greater than $t$. Under our stopping condition and general monotonicity of the logarithm this equates the probability $\mathbb{P}(\{S_t < \log (1/\delta)\})$, as the process has not stopped yet. Using above notation we may rewrite $S_t$ as $\sum_{i=1}^{t}v_i - A_i + \sum_{i=1}^{t}A_i$, yielding the equivalence
    $$\mathbb{P}(\{\tau > t\}) = \mathbb{P}(\{S_t < \log (1/\delta)\}) = \mathbb{P}\left(\left\{\frac{1}{t}\sum_{i=1}^{t}v_i - A_i + \frac{1}{t}\sum_{i=1}^{t}A_i < \frac{\log(1/\delta)}{t}\right\}\right).$$
    \noindent\emph{Martingale difference sequence.} It is clear that $(v_t- A_t)_{t \in \gT}$ forms a martingale difference sequence, and by boundedness conditions we have $|v_t - A_t| \leq \lambda_t$. Hence an application of Lemma~\ref{thm:azuma-hoeffding} will yield that $\mathbb{P}\left(\left\{|\frac{1}{t}\sum_{i=1}^{t}v_i - A_i| > \frac{\eta}{t}\right\}\right) \leq 2\exp\left(\frac{-\eta^{2}}{2\sum_{i=1}^{t}\lambda_i^{2}}\right)$ for some $\eta > 0$. We define this left-hand side event as $G_{t}^{c} = \left\{|\frac{1}{t}\sum_{i=1}^{t}v_i - A_i| > \frac{\eta}{t}\right\}$, an undesirable case where the martingale values tend to deviate overly far from its expectation. In contrast, the complementary event $G_t = \left\{|\frac{1}{t}\sum_{i=1}^{t}v_i - A_i| \leq \frac{\eta}{t}\right\}$ denotes the favourable case where difference values are strictly upper-bounded. Using a maximal (bounded) betting rate for which $\lambda_t \leq \lambda_{\text{max}} \; \forall t$ we have $\sum_{i=1}^{t}\lambda_{i}^{2} =  t\cdot\lambda_{\text{max}}^{2}$, and selecting $\eta = \sqrt{t \log t}$ we may then plug in values to claim that ${\mathbb{P}(G_{t}^{c}) \leq 2 \exp \left(-\log t/(2\lambda_{\text{max}}^{2})\right)}$, with a decaying rate in $t$. We can now rewrite the event probability $\mathbb{P}(\{\tau > t\})$ using the law of total probability as
    \begin{align*}
    \begin{split}
        \mathbb{P}(\{\tau > t\}) & \leq \mathbb{P}\left(\left\{\frac{1}{t}\sum_{i=1}^{t}A_i < \frac{\log(1/\delta)}{t} +|\frac{1}{t}\sum_{i=1}^{t}v_i - A_i|\right\} \cap G_{t}\right)  + \mathbb{P}(G_{t}^{c}) \\
        &\leq \mathbb{P}\left(\left\{\frac{1}{t}\sum_{i=1}^{t}A_i < \frac{\log(1/\delta)}{t} + \sqrt{\frac{\log t}{t}}\right\} \cap G_{t}\right)  + \mathbb{P}(G_{t}^{c}) \\
        & \leq \mathbb{P}\left(\left\{\frac{1}{t}\sum_{i=1}^{t}A_i < \frac{\log(1/\delta)}{t} + \sqrt{\frac{\log t}{t}}\right\}\right)  + \mathbb{P}(G_{t}^{c}).
    \end{split}
    \end{align*}
     Taking the limit $(\mathbb{P}\{\tau = \infty\}) \leq \limsup_{t \rightarrow \infty}\mathbb{P}(\{\tau > t\})$ we then bound the first term in the above expression and omit the second, \ie~$$\mathbb{P}(\{\tau = \infty\}) \leq \limsup_{t \rightarrow \infty}\mathbb{P}(\{\tau > t\}) \leq \mathbb{P}\left(\left\{\frac{1}{t}\sum_{i=1}^{t}A_i < \frac{\log(1/\delta)}{t} + \sqrt{\frac{\log t}{t}}\right\}\right).$$
    
    \noindent\emph{Under the alternative.} We are given that $\mathbb{E}[\rz_t - \epsilon \, | \, \gF_{t-1}] \leq 0$ for finitely many steps $t$, and $\mathbb{E}[\rz_t - \epsilon \, | \, \gF_{t-1}] > 0$ otherwise. Denote $\mu = \inf_{t \in \gT} \{\mathbb{E}_{H_1}[\rz_{t} - \epsilon \, | \, \mathcal{F}_{t-1}] > 0\}$ as the smallest risk violation among such time steps. Assuming a small, non-zero betting rate $\lambda_t$ we use the approximation $\log(1+x) \approx x$ to write $A_t = \mathbb{E}[\rv_t \, | \, \gF_{t-1}] = \mathbb{E}[\log (1 + \lambda_t \cdot \updelta_{t} \, | \, \mathcal{F}_{t-1}] \approx \lambda_{t} \cdot \mathbb{E}[\rz_t - \epsilon \, | \, \gF_{t-1}]$. By \emph{Cesàro means} we now have $\liminf_{t \rightarrow \infty} \frac{1}{t}\sum_{i=1}^{t}A_i \geq \liminf_{t \rightarrow \infty} A_t = \lambda \cdot \mu > 0$, where we use the above definition of $\mu$. Then for sufficiently large $t$ we have 
    $$\frac{1}{t}\sum_{i=1}^{t}A_i \gg \frac{\log (1/\delta)}{t} + \sqrt{\frac{\log t}{t}},$$
    and hence $\frac{1}{t}\sum_{i=1}^{t}A_i$ grows faster than the right-hand terms shrink, leading to the observation that in the limit $\limsup \mathbb{P}(\{\tau > t\}) \longrightarrow 0$ tends to zero. Thus under the alternative we have $\mathbb{P}_{H_1}(\{\tau = \infty\}) = 0$ and in complement $\mathbb{P}_{H_1}(\{\tau < \infty\}) = 1$, the desired \emph{power one property}.
\end{proof}

\subsection*{Details of Definition~\ref{thm:gro}.} 

We refer to the relevant works such as \cite{waudby2024estimating, shekhar2023near, grunwald2024safe} on the notion of \emph{growth rate optimality} and related betting rates. \cite{waudby2024estimating} also refer to the condition as \emph{growth rate adaptive to the particular alternative} (GRAPA), whereas \cite{grunwald2024safe} label it the \emph{GROW} criterion.

\subsection*{Proof of Proposition~\ref{thm:detection-delay-argument}.}
\label{app:math-proofs-detection-delay}

\noindent\textbf{Proposition~\ref{thm:detection-delay-argument}.}  (Detection delay bound). 
\emph{
    A worst-case detection delay for the hypothesis pair in \autoref{eq:hypotheses-formulation} and wealth process $M_t(\psi)$ in \autoref{eq:test-supermartingale} is characterized by ${(\tau(\psi) - \tau_*(\psi)) \approx \gO((\log(1/\delta) \, + \, T)/(\lambda \cdot \mu))}$, where $\mu$ denotes the risk violation intensity and $T$ a shift changepoint.
}
\begin{proof}
    We begin by clarifying the simplified problem setting underlying this result. We consider a simple risk violation scenario where, for any $\psi \in \gPsi$, an explicit time step $T \in \gT$ exists that induces a sufficiently strong shift. Thus we have $\mathbb{E}_{P_t}\left[\rz_t \ \vert  \ \mathcal{F}_{t-1}\right] \leq \epsilon$ for $t \leq T$, \ie, the risk is within tolerated bounds up to time $T$; and $\mathbb{E}_{P_t}\left[\rz_t \ \vert \ \mathcal{F}_{t-1}\right]> \epsilon$ for $t > T$, \ie, risk violation after time $T$. We formalize `sufficiently strong' by assuming that the risk deviates by a fixed term $\mu > 0$ from the boundary $\epsilon$, that is, $\mathbb{E}_{P_t}\left[\rz_t \ \vert \ \mathcal{F}_{t-1}\right] = \mu + \epsilon$ for $t > T$. This setting closely resembles a shift changepoint scenario, and akin to such works we desire to characterize the \emph{detection delay} $\tau(\psi) - \tau_*(\psi)$ and in particular the expected rates of delay in relation to different shift and risk monitoring parameters. 
    
    Let us once again start by considering the simpler summation wealth process (described in \autoref{sec:method} and \autoref{app:sum-process}), and suppress dependency on $\psi$ for notational clarity. For simplicity, consider a fixed betting rate $\lambda$ moving forward. Under the above shift setting, the process can be decomposed into its two temporal components for some $t > T$ as
    $$M_t = \sum_{i=1}^{t}\lambda \,(\rz_i - \epsilon) = \underbrace{\sum_{i=1}^{T}\lambda \,(\rz_i - \epsilon)}_{M_T} + \sum_{i=T+1}^{t}\lambda\left(\rz_i - \epsilon\right),$$ 
    where we abbreviate the first term to $M_T$. Expressing the pay-off term $(\rz_t - \epsilon)$ in terms of a decomposition as $\left(\rz_t - \mathbb{E}[\rz_t \, | \, \gF_{t-1}]\right) + \left(\mathbb{E}[\rz_t \, | \, \gF_{t-1}] - \epsilon\right)$, we rewrite $M_t$ as $$M_t = M_T + \sum_{i=T+1}^{t}\lambda\left(\rz_i - \mathbb{E}[\rz_i \, | \, \gF_{i-1}]\right) + \sum_{i=T+1}^{t}\lambda\underbrace{\left(\mathbb{E}[\rz_i \, | \, \gF_{i-1}] - \epsilon\right)}_{\mu},$$
    where by the assumption on a sufficient shift intensity the last term resolves to $(t-T) \cdot\lambda \cdot \mu$. Applying the same pay-off decomposition to $M_T$, we can regroup terms in the expression as
    $$M_t = \underbrace{\sum_{i=1}^{T}\lambda\left(\mathbb{E}[\rz_i \, | \, \gF_{i-1}] - \epsilon\right)}_{E_T} + \underbrace{\sum_{i=1}^{t}\lambda\left(\rz_i - \mathbb{E}[\rz_i \, | \, \gF_{i-1}]\right)}_{S_t} + \, (t-T) \cdot\lambda \cdot \mu,
    $$
    with abbreviated terms $E_T$ and $S_t$. We observe that $\left(\rz_t - \mathbb{E}[\rz_t \ \vert \ \mathcal{F}_{t-1}\right)_{t \in \gT}$ forms a martingale difference sequence, and hence by application of Lemma~\ref{thm:azuma-hoeffding} can be bounded. Employing the process' stopping time $\tau = \inf\{t \, : \, M_{t} \geq b\}$ for some predefined threshold $b>0$, we evaluate $M_t$ at $t=\tau$ to determine the detection delay to the actual changepoint $T$ as 
    \begin{align*}
    \begin{split}
        &E_T + S_\tau + (\tau - T) \cdot \lambda \cdot \mu = b, \\
        &\Leftrightarrow (\tau - T) \cdot \lambda \cdot \mu = b - E_T - S_{\tau}, \\
        &\Leftrightarrow \tau - T = \frac{b - E_T - S_\tau}{\lambda \cdot \mu}.
    \end{split}
    \end{align*}
    The \emph{expected} detection delay is then given by $\mathbb{E}[(\tau - T)] = \frac{b - E_T}{\lambda \cdot \mu}$, as $\mathbb{E}[S_{\tau}] = 0$. Since $E_{T} \leq 0$ surely, we have $E_{T} = 0$ in the best case (\ie~with maximum evidence) when $\mathbb{E}[\rz_t \ \vert \ \mathcal{F}_{t-1}] = \epsilon \; \forall t \leq T$, and $E_{T} = - T \cdot \lambda \cdot \epsilon$ in the worst case where  $\mathbb{E}[\rz_t \ \vert \ \mathcal{F}_{t-1}] = 0 \; \forall t \leq T$. This leads to a worst-case expected detection delay of $(\tau - T) \approx \gO(\frac{b + T}{\lambda \cdot \mu})$. To give a probabilistic statement we employ the one-sided case of Lemma~\ref{thm:azuma-hoeffding} on $S_t$, yielding the bound 
    $$\mathbb{P}\left(S_t  \geq \lambda \cdot \eta\right) \leq \exp\left(\frac{-\eta^{2}}{2t}\right),$$
    and for $\eta = \sqrt{t}/\lambda$ this equates $\mathbb{P}\left(S_t \geq \sqrt{t}\right) \leq \exp\left(-\frac{1}{2\lambda^{2}}\right)$. We then have the worst-case delay bounded as ${(\tau - T) = \frac{b + T \cdot \lambda \cdot \epsilon - S_{\tau}}{\lambda \cdot \mu} \leq \frac{b + T \cdot \lambda \cdot \epsilon - \sqrt{\tau}}{\lambda \cdot \mu}}$, which further results in $(\tau - T) \approx \gO\left(\frac{b + T \cdot \lambda \cdot \epsilon}{\lambda \cdot \mu}\right)$ with high probability.
    
    The same arguments can be extended to the multiplicative wealth process $M_t = \prod_{i=1}^{t}\left(1 + \lambda_i \left(\rz_i - \epsilon\right)\right)$ (\autoref{eq:test-supermartingale}) by considering $\log M_t$ and using the approximation $\log (1+x) \approx x$, valid for sufficiently small bets $\lambda$. The log-wealth process $\log M_t \approx \sum_{i=1}^{t}\lambda \left(\rz_i - \epsilon\right)$ then equates the summation process considered above. For the final result, we instantiate $b = \log (1/\delta)$ to obtain the characterization $(\tau - T) \approx \gO\left(\frac{\log (1/\delta) \,+\, T}{\lambda \cdot \mu}\right)$. Some intituitive insights from the expression are that \emph{(i)} the detection delay is directly proportional to $T$, the time step at which risk violation begins---if this shift changepoint starts later, then additional delay arises from having to overcome the initial martingale's decay; and \emph{(ii)} the detection delay is inversely proportional to both the betting rate $\lambda$ and the intensity of the violations $\mu$. However, we note that our methodology comes with the flexibility to control these delays to some extent by leveraging a smart betting rate design.
\end{proof}

\section{Additional Related Work}
\label{app:background}

\paragraph{Static risk control.} The framework of \emph{conformal prediction} constructs set predictors with upper bounds specifically on the miscoverage risk under \emph{i.i.d.} or exchangeable data, with a substantial recent body of literature (see, \eg, \cite{angelopoulos2023conformal, fontana2023conformal}). The approach has been extended to more general bounded risks by \cite{angelopoulos2024crc} leveraging similar exchangeability arguments. \cite{bates2021distribution} use concentration inequalities to provide high-probability assurances for monotonic expectation risks, and \cite{angelopoulos2021learn} extend the idea to non-monotonic risks by reframing the task as a non-sequential multiple testing problem. Related in spirit, \cite{angelopoulos2023prediction} propose the use of a hold-out unlabelled dataset to provide probability guarantees for confidence intervals on population-level parameters. 

\paragraph{Risk control for stream data and under shift.} Recent work on conformal prediction includes addressing non-exchangeable data sequences such as time series, \eg~by tracking and updating the tolerated miscoverage rate \citep{Gibbs2021AdaptiveCI, angelopoulos2024online, zaffran2022adaptive} or different weighting schemes \citep{barber2023conformal, guan2023localized}. Particular applications also include outlier detection \citep{Bates2021TestingFO, laxhammar2015inductive}. Recent work on data shifts has included covariate shift \citep{tibshirani2019conformal}, label shift \citep{podkopaev2021distribution} and their abstraction to more general shifts \citep{prinster2024conformal}. All of the above work predominantly focuses on the miscoverage risk, with \cite{Feldman2022AchievingRC} being an interesting extension of \cite{Gibbs2021AdaptiveCI} to more general bounded risks, discussed in \autoref{subsec:connection-methods}. Furthermore, obtainable guarantees are generally asymptotic or finite-sample only under relaxation (\eg, with respect to a permitted coverage deviation from the targeted guarantee).

\section{Additional Experimental Design}
\label{app:sec-exp-design}

\paragraph{Empirical-Bernstein wealth process.} We directly adopt the formulation as a test supermartingale or wealth process described in \cite{waudby2024estimating} (see Sec. 3.2 and Thm. 2 in their paper), given by
\begin{equation*}
    M^{EB}_{t}(\psi) = \prod_{i=1}^{t} \exp\{ \lambda_i \, (z_i - \epsilon) - v_i \, \rho(\lambda_i) \}
\end{equation*}
with $v_i = 4\,(z_i - \hat{\mu}_{i-1})^2$ and $\rho(\lambda_i) = 1/4\,(- \log(1 - \lambda_i) - \lambda_i)$ for $\lambda_i \in [0,1)$, and using the suggested \emph{predictable plug-in} betting rate $\lambda^{EB}_i = \min\{\sqrt{\frac{2 \, \log(2/\delta)}{\hat{\sigma}^2_{i-1} \, i \, \log(1 + i)}}, c\}$. The estimates $\hat{\mu}_{i-1}$ and $\hat{\sigma}^2_{i-1}$ denote the empirical running mean and variance over the observed loss sequence $\{z_1, \dots, z_{i-1}\}$ up to time $i-1$, thus rendering $\lambda^{EB}_i$ predictable at every time step. We select $c = 1/2$ as a recommended constant $c \in (0,1)$, and omit the bias terms of $1/4$ and $1/2$ for $\hat{\mu}_{i-1}$ and $\hat{\sigma}^2_{i-1}$ respectively, which are negligable for sufficiently large streams. \cite{podkopaev2021tracking} leverage the process in its confidence sequence-equivalent form to estimate bounds on the running risk $R_r(\psi)$ in their problem setting, motivating its inclusion as a baseline.

\paragraph{Choice of betting rate.} We refer to \cite{waudby2024estimating} on the particular technical details of various betting rate designs, in particular their App. B.2 for GRO and App. B.3 for \emph{approximately GRO}. In essence, a direct optimization of the GRO condition (Definition~\ref{thm:gro}) can be achieved by exhaustive root-finding over a fine grid of possible values for $\lambda_t \in [0, 1/\epsilon)$. The approximation to GRO takes an additional Taylor approximation and truncation step to derive a closed-form solution for the root, thus being substantially more efficient. Our final betting rate expression takes this approach for a particular truncation aligned with our problem setting (\ie~the permitted range of $\lambda_t$). Other approximations to the GRO objective are also possible, such as solving for a lower-bound to the wealth (see their App. B.4 and following).

\paragraph{Batched data stream.} Assume we observe more than a single observation at every time $t$, \ie~the data stream with samples $\{ (\vx_{t,b}, \vy_{t,b})\}_{b=1}^{B} \sim P_t$ for some batch size $B \ll B_*$. Then we can average over the evidence in each batch to obtain a more robust measure of evidence for risk violations as
\begin{equation*}
    M_t(\psi) = \prod_{i=1}^{t} \frac{1}{B} \sum_{b=1}^{B} (1 + \lambda_{i}(z_{i,b} - \epsilon)),
\end{equation*}
leading to a reduced variance of the wealth process as well as reducing the detection delay $\tau(\psi) - \tau_*(\psi)$ with respect to the true risk. However this does not necessarily equate lower sampling costs, since the total number of observations is $B \cdot t$.

\paragraph{False alarm rate.} For a given experiment run or trial, a threshold candidate $\psi$ raises a false alarm (or is labelled a false positive) if $\gR_t(\psi) \leq \epsilon$ but $M_t(\psi) \geq 1/\delta$, thus erroneously claiming violation. Equivalently, we may state for the detection delay that $(\tau(\psi) - \tau_*(\psi)) < 0$ stops prematurely. The false alarm rate is then computed as the fraction of false alarms across $R$ trials, \ie
\begin{equation*}
\label{eq:false-alarm-rate}
    \%FP = \frac{1}{R}\sum_{r=1}^{R} \mathbbm{1}[(\tau(\psi) - \tau_*(\psi)) < 0],
\end{equation*}
and compared to the tolerated false alarm rate $\delta \in (0,1)$. If $\%FP > \delta$ then the Type-I error under $H_0(\psi)$ is uncontrolled, resp. any error control property violated.

\paragraph{Total error rate (TER).} In \autoref{subsec:exp-ood} the target risk to monitor is the \emph{total error rate} (TER), accounting for both cases of inlier (false positives, FP) and outlier misclassification (false negatives, FN). That is, we define the true risk $\gR_t(\psi) = \mathbb{E}_{P_t}[\rz_t]$ with loss variable
\begin{equation*}
    \rz_t = 
    \begin{cases}
        1, & \text{if } \texttt{out}(\rvx_t) \geq \psi \; \text{ and }\; (\rvx_t, \rvy_t) \sim P_{in}, \hfill \text{ (FP)} \\
        1, & \text{if } \texttt{out}(\rvx_t) < \psi \; \text{ and }\; (\rvx_t, \rvy_t) \sim P_{out}, \; \text{ (FN)} \\
        0 & \text{else.}
    \end{cases}
\end{equation*}
The TER is a complex risk quantity that is both non-monotonic across time \emph{and} thresholds, since the FP and FN terms introduce competing objectives in terms of what constitutes a `safe' threshold $\psi$. Under a stepwise shift with increasing outlier fraction, the FP term initially weighs stronger, motivating a higher threshold choice (since the chance of an inlier mislabelling is reduced). However, as the outlier fraction increases the FN term becomes more relevant, motivating a lower threshold choice (\ie~increasing the chance of an outlier label). Therefore, no clear `trivial' safe threshold selection is available except for $\hat{\psi} = 1$ if $\pi_{t}^{out} = 0$ and $\hat{\psi} = 0$ if $\pi_{t}^{out} = 1$.

\paragraph{Miscoverage rate (MCR).} In \autoref{subsec:exp-sets} the target risk to monitor is the \emph{miscoverage rate}, accounting for set exclusion of the correct label $\vy_t$. That is, we define the true risk $\gR_t(\psi) = \mathbb{E}_{P_t}[\rz_t]$ with loss variable $\rz_t = \mathbbm{1}[\vy_t \notin \hat{f}_{\psi}(\rvx_t)]$. In this case, the MCR is closer to monotonically increasing over time as the natural temporal shift caused by both FMoW and Naval propulsion degrades model performance, and clearly monotonic in the thresholds. For FMoW, an indefinitely valid `safe' threshold with zero risk is given by $\hat{\psi} = 0$, resulting in a prediction set matching the full label space $\gY$, and thus $\rz_t = 0$ at every time step. For Naval propulsion, since $\gY \subseteq [0,1]$ and in practice $\vy_t \in [0.95, 1.0]$ we instantiate a fine grid of threshold candidates in the range $\gPsi := [0, 0.05]$. Thus for a sufficiently well-trained regressor, $\hat{\psi} = 0.05$ trivially ensures coverage (again returning the full response space) while thresholds towards zero place higher reliance on the prediction's accuracy at the risk of miscoverage. Clearly, such `trival' threshold solutions are generally impractical, but the MCR's monotonic behaviour renders it a more interpretable quantity and easier to track than the TER in \autoref{subsec:exp-ood}. 

\paragraph{Functional Map of the World dataset.} We consider the \emph{Functional Map of the World} dataset (FMoW) \citep{christie2018functional}, a large-scale satellite image dataset with 62 categories of building and land use, collected across various geographic regions and over 16 years (2002 -- 2017). We consider the time-dependent partitioning of FMoW proposed by \cite{yao2022wild}, wherein a natural shift occurs as land use for the \emph{same} satellite image locations, surveyed repeatedly over several years, changes over time. The predictor (a DenseNet-121) is trained on earlier years (2002 -- 2012), and we increase the test stream frequency to simulate daily observations by sampling chronologically from data in 2013 -- 2017 every 365 time steps (equating each passing year). This induces a (slow) step-wise shift, and we observe that the classifier's predictive accuracy worsens as time progresses, in line with results reported by \cite{yao2022wild}.

\paragraph{Naval propulsion system dataset.} We consider predictive maintenance (or equipment monitoring) data on naval gas turbine behaviour \citep{cipollini2018condition}. This tabular time series consists of $\sim 12 \, 000$ recordings for various turbine system parameters, and an associated turbine compressor degradation coefficient denoting the compressor’s health. Over time this degradation coefficient steadily increases from 0.95 to 1.0, denoting a gradual equipment decay. We supplement the data via jittered resampling of early observations (the first 2000 samples) to enrich the initial `healthy' compressor state, and train a Random Forest regressor on that data. Expectedly, as the compressor gradually degrades beyond its initial `healthy' range the predictor fails to extrapolate, resulting in decreased performance in line with a temporal distribution shift.

\newpage

\section{Additional Experimental Results}
\label{app:sec-exp-res}

\begin{figure*}[!h]
    \centering
    \includegraphics[width=1\linewidth]{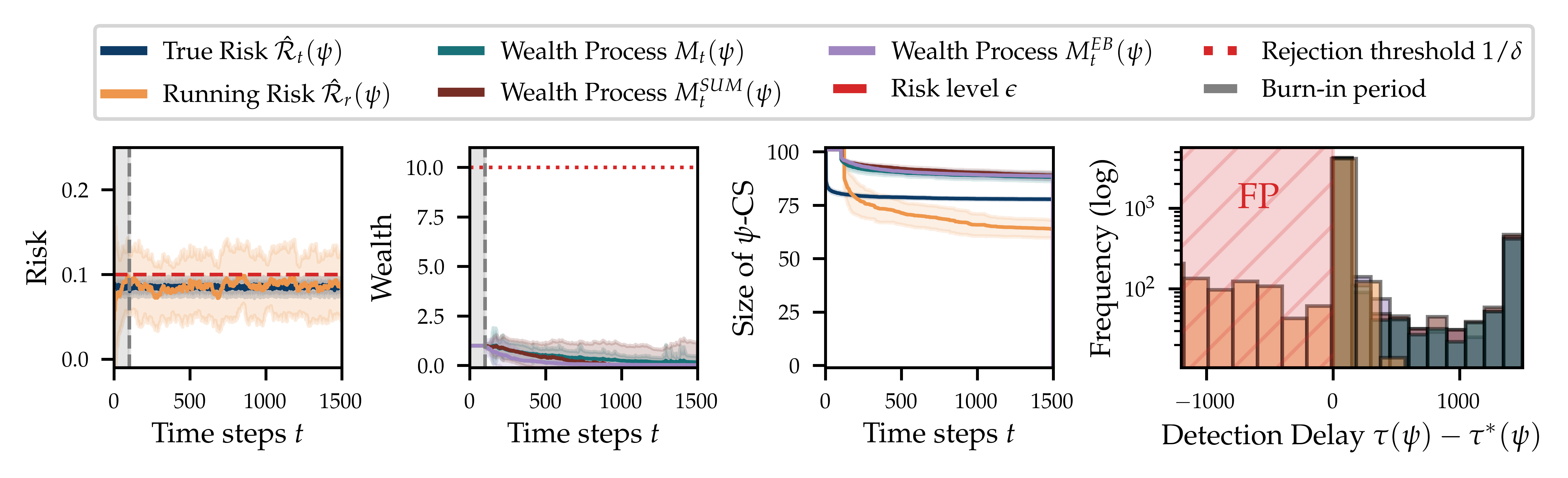}
    \caption{
    Results for \textbf{outlier detection with no shift} (\autoref{subsec:exp-ood}). \emph{From left to right:} Visuals of the steady risk and wealth process behaviour with respective rejection thresholds $\epsilon$ and $1/\delta$, for a single threshold candidate (here $\psi=0.20$); the behaviour of the valid threshold set $\psi$-CS (\autoref{eq:psi-cs}), which slightly shrinks by eliminating clearly violating thresholds and then stabilizes, signalling robust detection performance for the \emph{i.i.d.} stream; and the empirical distributions of detection delays $\tau(\psi) - \tau_*(\psi)$ across all $\psi \in \gPsi$, including the false alarm region (FP). We also have $B=1$ and $S=50$, with results evaluated over $R=50$ trials (mean and std. deviation).
    }
    \label{fig:app-ood-noshift}
\end{figure*}

\begin{figure*}[!h]
    \centering
    \includegraphics[width=1\linewidth]{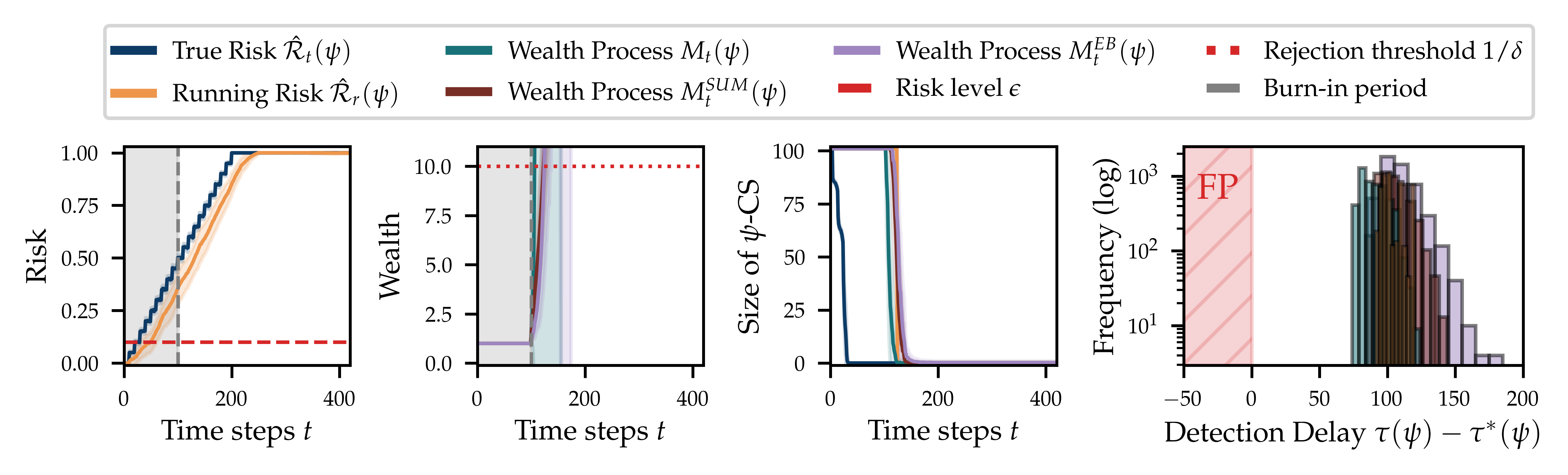}
    \caption{
    Results for \textbf{outlier detection with an immediate shift early on} (\autoref{subsec:exp-ood}). \emph{From left to right:} Visuals of the strongly growing risk and wealth process behaviour with respective rejection thresholds $\epsilon$ and $1/\delta$, for a single threshold candidate (here $\psi=0.90$); the behaviour of the valid threshold set $\psi$-CS (\autoref{eq:psi-cs}), which almost immediately collapses to zero signalling a highly unreliable model in need of updating; and the empirical distributions of detection delays $\tau(\psi) - \tau_*(\psi)$ across all $\psi \in \gPsi$, including the false alarm region (FP). Since the evidence for risk violation is very clear, no false alarms are incurred by any tracker. We also have $B=1$ and $S=50$, with results evaluated over $R=50$ trials (mean and std. deviation).
    }
    \label{fig:app-ood-direct}
\end{figure*}

\begin{table*}[!ht]
    \caption{
    Results for \textbf{outlier detection with a stepwise shift} (\autoref{subsec:exp-ood}). We provide monitoring results of the total error rate for $\epsilon=0.1, \delta=0.1$ and different combinations of the two key tracking parameters, sliding window size $S$ and batch size $B$. The delay quantity $\bar{\tau}(\psi)$ denotes the mean and std. deviation of detection delays $\tau(\psi) - \tau_*(\psi)$ across repeated trials ($R=50$), whereas $\%FP > 0$ and $\%FP > \delta$ denote the number of thresholds $\psi \in \gPsi$ (in \%) whose false alarm rate (see \autoref{eq:false-alarm-rate}) is non-zero or exceeds the desired rate $\delta$, respectively.
    }
    \resizebox{\linewidth}{!}{
    \begin{tabular}{llrrrrrrrrrrrr}
        \toprule
        \multicolumn{2}{c}{\textbf{Params}} & \multicolumn{3}{c}{\textbf{Running Risk $\hat{\gR}_r(\psi)$}} & \multicolumn{3}{c}{\textbf{Wealth Process $M_t(\psi)$}} & \multicolumn{3}{c}{\textbf{Wealth Process $M^{SUM}_t(\psi)$}} & \multicolumn{3}{c}{\textbf{Wealth Process $M^{EB}_t(\psi)$}} \\
        \cmidrule(lr){3-5} \cmidrule(lr){6-8} \cmidrule(lr){9-11} \cmidrule(lr){12-14}
        \textbf{$S$} & \textbf{$B$} & Delay $\bar{\tau}(\psi)$ & \%FP $> 0$ & \%FP $> \delta$ & Delay $\bar{\tau}(\psi)$ & \%FP $> 0$ & \%FP $> \delta$ & Delay $\bar{\tau}(\psi)$ & \%FP $> 0$ & FP $> \delta$ & Delay $\bar{\tau}(\psi)$ & \%FP $> 0$ & \%FP $> \delta$ \\
        \midrule
        None & 1 & $408 \pm \scriptstyle 201$ & 13.86\% & 3.96\% & $563 \pm \scriptstyle 207$ & 0.00\% & 0.00\% & $694 \pm \scriptstyle 230$ & 0.00\% & 0.00\% & $807 \pm \scriptstyle 334$ & 0.00\% & 0.00\% \\
        None & 10 & $393 \pm \scriptstyle 214$ & 6.93\% & 2.97\% & $441 \pm \scriptstyle 210$ & 0.00\% & 0.00\% & $495 \pm \scriptstyle 219$ & 0.00\% & 0.00\% & $708 \pm \scriptstyle 307$ & 0.00\% & 0.00\% \\
        None & 50 & $393 \pm \scriptstyle 213$ & 0.99\% & 0.00\% & $414 \pm \scriptstyle 215$ & 0.00\% & 0.00\% & $451 \pm \scriptstyle 220$ & 0.00\% & 0.00\% & $655 \pm \scriptstyle 273$ & 0.00\% & 0.00\% \\
        \midrule
        200 & 1 & $180 \pm \scriptstyle 84$ & 25.74\% & 3.96\% & $355 \pm \scriptstyle 110$ & 0.00\% & 0.00\% & $469 \pm \scriptstyle 128$ & 0.00\% & 0.00\% & $831 \pm \scriptstyle 337$ & 0.00\% & 0.00\% \\
        200 & 10 & $159 \pm \scriptstyle 74$ & 6.93\% & 2.97\% & $210 \pm \scriptstyle 80$ & 0.00\% & 0.00\% & $257 \pm \scriptstyle 92$ & 0.00\% & 0.00\% & $698 \pm \scriptstyle 301$ & 0.00\% & 0.00\% \\
        200 & 50 & $156 \pm \scriptstyle 70$ & 0.99\% & 0.00\% & $171 \pm \scriptstyle 73$ & 0.00\% & 0.00\% & $200 \pm \scriptstyle 75$ & 0.00\% & 0.00\% & $604 \pm \scriptstyle 240$ & 0.00\% & 0.00\% \\
        \midrule
        50 & 1 & $77 \pm \scriptstyle 83$ & 79.21\% & 71.29\% & $349 \pm \scriptstyle 123$ & 0.00\% & 0.00\% & $444 \pm \scriptstyle 127$ & 0.00\% & 0.00\% & $833 \pm \scriptstyle 333$ & 0.00\% & 0.00\% \\
        50 & 10 & $68 \pm \scriptstyle 46$ & 15.84\% & 3.96\% & $176 \pm \scriptstyle 83$ & 0.00\% & 0.00\% & $214 \pm \scriptstyle 79$ & 0.00\% & 0.00\% & $695 \pm \scriptstyle 300$ & 0.00\% & 0.00\% \\
        50 & 50 & $74 \pm \scriptstyle 50$ & 0.99\% & 0.00\% & $131 \pm \scriptstyle 81$ & 0.00\% & 0.00\% & $167 \pm \scriptstyle 77$ & 0.00\% & 0.00\% & $598 \pm \scriptstyle 238$ & 0.00\% & 0.00\% \\
        \midrule
        10 & 1 & $37 \pm \scriptstyle 79$ & 80.20\% & 73.27\% & $394 \pm \scriptstyle 151$ & 0.00\% & 0.00\% & $465 \pm \scriptstyle 139$ & 0.00\% & 0.00\% & $865 \pm \scriptstyle 326$ & 0.00\% & 0.00\% \\
        10 & 10 & $18 \pm \scriptstyle 42$ & 75.25\% & 27.72\% & $184 \pm \scriptstyle 93$ & 0.00\% & 0.00\% & $204 \pm \scriptstyle 78$ & 0.00\% & 0.00\% & $685 \pm \scriptstyle 292$ & 0.00\% & 0.00\% \\
        10 & 50 & $41 \pm \scriptstyle 29$ & 5.94\% & 0.00\% & $120 \pm \scriptstyle 86$ & 0.00\% & 0.00\% & $160 \pm \scriptstyle 78$ & 0.00\% & 0.00\% & $598 \pm \scriptstyle 239$ & 0.00\% & 0.00\% \\
        \bottomrule
    \end{tabular}
    }
    \label{tab:app-ood-step}
\end{table*}

\begin{table*}[!ht]
    \caption{
    Results for \textbf{outlier detection with no shift} (\autoref{subsec:exp-ood}). We provide monitoring results of the total error rate for $\epsilon=0.1, \delta=0.1$ and different combinations of the two key tracking parameters, sliding window size $S$ and batch size $B$. The delay quantity $\bar{\tau}(\psi)$ denotes the mean and std. deviation of detection delays $\tau(\psi) - \tau_*(\psi)$ across repeated trials ($R=50$), whereas $\%FP > 0$ and $\%FP > \delta$ denote the number of thresholds $\psi \in \gPsi$ (in \%) whose false alarm rate (see \autoref{eq:false-alarm-rate}) is non-zero or exceeds the desired rate $\delta$, respectively.
    }
    \resizebox{\linewidth}{!}{
    \begin{tabular}{llrrrrrrrrrrrr}
        \toprule
        \multicolumn{2}{c}{\textbf{Params}} & \multicolumn{3}{c}{\textbf{Running Risk $\hat{\gR}_r(\psi)$}} & \multicolumn{3}{c}{\textbf{Wealth Process $M_t(\psi)$}} & \multicolumn{3}{c}{\textbf{Wealth Process $M^{SUM}_t(\psi)$}} & \multicolumn{3}{c}{\textbf{Wealth Process $M^{EB}_t(\psi)$}} \\
        \cmidrule(lr){3-5} \cmidrule(lr){6-8} \cmidrule(lr){9-11} \cmidrule(lr){12-14}
        \textbf{$S$} & \textbf{$B$} & Delay $\bar{\tau}(\psi)$ & \%FP $> 0$ & \%FP $> \delta$ & Delay $\bar{\tau}(\psi)$ & \%FP $> 0$ & \%FP $> \delta$ & Delay $\bar{\tau}(\psi)$ & \%FP $> 0$ & \%FP $> \delta$ & Delay $\bar{\tau}(\psi)$ & \%FP $> 0$ & \%FP $> \delta$ \\
        \midrule
        None & 1 & $53 \pm \scriptstyle 277$ & 17.82\% & 4.95\% & $176 \pm \scriptstyle 437$ & 0.00\% & 0.00\% & $197 \pm \scriptstyle 463$ & 0.00\% & 0.00\% & $185 \pm \scriptstyle 446$ & 0.00\% & 0.00\% \\
        None & 10 & $70 \pm \scriptstyle 289$ & 6.93\% & 0.00\% & $127 \pm \scriptstyle 389$ & 0.00\% & 0.00\% & $137 \pm \scriptstyle 398$ & 0.00\% & 0.00\% & $176 \pm \scriptstyle 451$ & 0.00\% & 0.00\% \\
        None & 50 & $84 \pm \scriptstyle 319$ & 2.97\% & 0.00\% & $112 \pm \scriptstyle 373$ & 0.00\% & 0.00\% & $118 \pm \scriptstyle 376$ & 0.00\% & 0.00\% & $180 \pm \scriptstyle 463$ & 0.00\% & 0.00\% \\
        \midrule
        200 & 1 & $11 \pm \scriptstyle 233$ & 17.82\% & 11.88\% & $175 \pm \scriptstyle 434$ & 0.00\% & 0.00\% & $197 \pm \scriptstyle 462$ & 0.00\% & 0.00\% & $185 \pm \scriptstyle 445$ & 0.00\% & 0.00\% \\
        200 & 10 & $56 \pm \scriptstyle 243$ & 6.93\% & 0.00\% & $127 \pm \scriptstyle 389$ & 0.00\% & 0.00\% & $137 \pm \scriptstyle 398$ & 0.00\% & 0.00\% & $176 \pm \scriptstyle 451$ & 0.00\% & 0.00\% \\
        200 & 50 & $80 \pm \scriptstyle 304$ & 2.97\% & 0.00\% & $111 \pm \scriptstyle 372$ & 0.00\% & 0.00\% & $118 \pm \scriptstyle 376$ & 0.00\% & 0.00\% & $180 \pm \scriptstyle 463$ & 0.00\% & 0.00\% \\
        \midrule
        50 & 1 & $-102 \pm \scriptstyle 362$ & 27.72\% & 21.78\% & $177 \pm \scriptstyle 437$ & 0.00\% & 0.00\% & $197 \pm \scriptstyle 461$ & 0.00\% & 0.00\% & $185 \pm \scriptstyle 446$ & 0.00\% & 0.00\% \\
        50 & 10 & $13 \pm \scriptstyle 107$ & 8.91\% & 4.95\% & $127 \pm \scriptstyle 390$ & 0.00\% & 0.00\% & $135 \pm \scriptstyle 395$ & 0.00\% & 0.00\% & $177 \pm \scriptstyle 452$ & 0.00\% & 0.00\% \\
        50 & 50 & $59 \pm \scriptstyle 245$ & 2.97\% & 0.00\% & $112 \pm \scriptstyle 371$ & 0.00\% & 0.00\% & $118 \pm \scriptstyle 375$ & 0.00\% & 0.00\% & $180 \pm \scriptstyle 463$ & 0.00\% & 0.00\% \\
        \midrule
        10 & 1 & $-226 \pm \scriptstyle 478$ & 37.62\% & 33.66\% & $183 \pm \scriptstyle 447$ & 0.00\% & 0.00\% & $198 \pm \scriptstyle 461$ & 0.00\% & 0.00\% & $199 \pm \scriptstyle 470$ & 0.00\% & 0.00\% \\
        10 & 10 & $-126 \pm \scriptstyle 363$ & 19.80\% & 17.82\% & $133 \pm \scriptstyle 402$ & 0.00\% & 0.00\% & $129 \pm \scriptstyle 382$ & 0.00\% & 0.00\% & $178 \pm \scriptstyle 455$ & 0.00\% & 0.00\% \\
        10 & 50 & $12 \pm \scriptstyle 73$ & 7.92\% & 2.97\% & $111 \pm \scriptstyle 370$ & 0.00\% & 0.00\% & $118 \pm \scriptstyle 373$ & 0.00\% & 0.00\% & $180 \pm \scriptstyle 463$ & 0.00\% & 0.00\% \\
        \bottomrule
    \end{tabular}
    }
    \label{tab:app-ood-noshift}
\end{table*}

\begin{table*}[!ht]
    \caption{
    Results for \textbf{outlier detection with an immediate shift early on} (\autoref{subsec:exp-ood}). We provide monitoring results of the total error rate for $\epsilon=0.1, \delta=0.1$ and different combinations of the two key tracking parameters, sliding window size $S$ and batch size $B$. The delay quantity $\bar{\tau}(\psi)$ denotes the mean and std. deviation of detection delays $\tau(\psi) - \tau_*(\psi)$ across repeated trials ($R=50$), whereas $\%FP > 0$ and $\%FP > \delta$ denote the number of thresholds $\psi \in \gPsi$ (in \%) whose false alarm rate (see \autoref{eq:false-alarm-rate}) is non-zero or exceeds the desired rate $\delta$, respectively.
    }
    \resizebox{\linewidth}{!}{
    \begin{tabular}{llrrrrrrrrrrrr}
        \toprule
        \multicolumn{2}{c}{\textbf{Params}} & \multicolumn{3}{c}{\textbf{Running Risk $\hat{\gR}_r(\psi)$}} & \multicolumn{3}{c}{\textbf{Wealth Process $M_t(\psi)$}} & \multicolumn{3}{c}{\textbf{Wealth Process $M^{SUM}_t(\psi)$}} & \multicolumn{3}{c}{\textbf{Wealth Process $M^{EB}_t(\psi)$}} \\
        \cmidrule(lr){3-5} \cmidrule(lr){6-8} \cmidrule(lr){9-11} \cmidrule(lr){12-14}
        \textbf{$S$} & \textbf{$B$} & Delay $\bar{\tau}(\psi)$ & \%FP $> 0$ & \%FP $> \delta$ & Delay $\bar{\tau}(\psi)$ & \%FP $> 0$ & \%FP $> \delta$ & Delay $\bar{\tau}(\psi)$ & \%FP $> 0$ & \%FP $> \delta$ & Delay $\bar{\tau}(\psi)$ & \%FP $> 0$ & \%FP $> \delta$ \\
        \midrule
        None & 1 & $104 \pm \scriptstyle 9$ & 0.00\% & 0.00\% & $91 \pm \scriptstyle 9$ & 0.00\% & 0.00\% & $108 \pm \scriptstyle 9$ & 0.00\% & 0.00\% & $110 \pm \scriptstyle 79$ & 0.00\% & 0.00\% \\
        None & 10 & $38 \pm \scriptstyle 9$ & 0.00\% & 0.00\% & $28 \pm \scriptstyle 10$ & 0.00\% & 0.00\% & $40 \pm \scriptstyle 12$ & 0.00\% & 0.00\% & $49 \pm \scriptstyle 11$ & 0.00\% & 0.00\% \\
        None & 50 & $36 \pm \scriptstyle 9$ & 0.00\% & 0.00\% & $23 \pm \scriptstyle 11$ & 0.00\% & 0.00\% & $32 \pm \scriptstyle 11$ & 0.00\% & 0.00\% & $50 \pm \scriptstyle 13$ & 0.00\% & 0.00\% \\
        \midrule
        200 & 1 & $104 \pm \scriptstyle 9$ & 0.00\% & 0.00\% & $91 \pm \scriptstyle 9$ & 0.00\% & 0.00\% & $108 \pm \scriptstyle 9$ & 0.00\% & 0.00\% & $110 \pm \scriptstyle 79$ & 0.00\% & 0.00\% \\
        200 & 10 & $38 \pm \scriptstyle 9$ & 0.00\% & 0.00\% & $28 \pm \scriptstyle 10$ & 0.00\% & 0.00\% & $40 \pm \scriptstyle 12$ & 0.00\% & 0.00\% & $49 \pm \scriptstyle 11$ & 0.00\% & 0.00\% \\
        200 & 50 & $36 \pm \scriptstyle 9$ & 0.00\% & 0.00\% & $23 \pm \scriptstyle 11$ & 0.00\% & 0.00\% & $32 \pm \scriptstyle 11$ & 0.00\% & 0.00\% & $50 \pm \scriptstyle 13$ & 0.00\% & 0.00\% \\
        \midrule
        50 & 1 & $104 \pm \scriptstyle 9$ & 0.00\% & 0.00\% & $90 \pm \scriptstyle 9$ & 0.00\% & 0.00\% & $105 \pm \scriptstyle 40$ & 0.00\% & 0.00\% & $110 \pm \scriptstyle 57$ & 0.00\% & 0.00\% \\
        50 & 10 & $38 \pm \scriptstyle 9$ & 0.00\% & 0.00\% & $27 \pm \scriptstyle 9$ & 0.00\% & 0.00\% & $37 \pm \scriptstyle 9$ & 0.00\% & 0.00\% & $48 \pm \scriptstyle 11$ & 0.00\% & 0.00\% \\
        50 & 50 & $36 \pm \scriptstyle 9$ & 0.00\% & 0.00\% & $22 \pm \scriptstyle 10$ & 0.00\% & 0.00\% & $30 \pm \scriptstyle 9$ & 0.00\% & 0.00\% & $50 \pm \scriptstyle 13$ & 0.00\% & 0.00\% \\
        \midrule
        10 & 1 & $104 \pm \scriptstyle 9$ & 0.00\% & 0.00\% & $90 \pm \scriptstyle 10$ & 0.00\% & 0.00\% & $110 \pm \scriptstyle 88$ & 0.00\% & 0.00\% & $110 \pm \scriptstyle 57$ & 0.00\% & 0.00\% \\
        10 & 10 & $26 \pm \scriptstyle 5$ & 0.00\% & 0.00\% & $18 \pm \scriptstyle 6$ & 0.00\% & 0.00\% & $27 \pm \scriptstyle 6$ & 0.00\% & 0.00\% & $46 \pm \scriptstyle 11$ & 0.00\% & 0.00\% \\
        10 & 50 & $25 \pm \scriptstyle 3$ & 0.00\% & 0.00\% & $11 \pm \scriptstyle 4$ & 0.00\% & 0.00\% & $22 \pm \scriptstyle 5$ & 0.00\% & 0.00\% & $48 \pm \scriptstyle 12$ & 0.00\% & 0.00\% \\
        \bottomrule
    \end{tabular}
    }
    \label{tab:app-ood-direct}
\end{table*}

\begin{table*}[!ht]
    \caption{
    Results for \textbf{set prediction with a temporal shift on FMoW} (\autoref{subsec:exp-sets}). We provide monitoring results of the miscoverage rate for $\epsilon=0.1, \delta=0.1$ and different combinations of the two key tracking parameters, sliding window size $S$ and batch size $B$. The delay quantity $\bar{\tau}(\psi)$ denotes the mean and std. deviation of detection delays $\tau(\psi) - \tau_*(\psi)$ across repeated trials ($R=50$), whereas $\%FP > 0$ and $\%FP > \delta$ denote the number of thresholds $\psi \in \gPsi$ (in \%) whose false alarm rate (see \autoref{eq:false-alarm-rate}) is non-zero or exceeds the desired rate $\delta$, respectively.
    }
    \resizebox{\linewidth}{!}{
    \begin{tabular}{llrrrrrrrrrrrr}
        \toprule
        \multicolumn{2}{c}{\textbf{Params}} & \multicolumn{3}{c}{\textbf{Running Risk $\hat{\gR}_r(\psi)$}} & \multicolumn{3}{c}{\textbf{Wealth Process $M_t(\psi)$}} & \multicolumn{3}{c}{\textbf{Wealth Process $M^{SUM}_t(\psi)$}} & \multicolumn{3}{c}{\textbf{Wealth Process $M^{EB}_t(\psi)$}} \\
        \cmidrule(lr){3-5} \cmidrule(lr){6-8} \cmidrule(lr){9-11} \cmidrule(lr){12-14}
        \textbf{$S$} & \textbf{$B$} & Delay $\bar{\tau}(\psi)$ & \%FP $> 0$ & \%FP $> \delta$ & Delay $\bar{\tau}(\psi)$ & \%FP $> 0$ & \%FP $> \delta$ & Delay $\bar{\tau}(\psi)$ & \%FP $> 0$ & \%FP $> \delta$ & Delay $\bar{\tau}(\psi)$ & \%FP $> 0$ & \%FP $> \delta$ \\
        \midrule
        None & 1 & $166 \pm \scriptstyle 192$ & 3.96\% & 1.98\% & $262 \pm \scriptstyle 313$ & 0.00\% & 0.00\% & $340 \pm \scriptstyle 359$ & 0.00\% & 0.00\% & $324 \pm \scriptstyle 351$ & 0.00\% & 0.00\% \\
        None & 10 & $82 \pm \scriptstyle 196$ & 1.98\% & 0.00\% & $103 \pm \scriptstyle 263$ & 0.00\% & 0.00\% & $126 \pm \scriptstyle 281$ & 0.00\% & 0.00\% & $213 \pm \scriptstyle 380$ & 0.00\% & 0.00\% \\
        None & 50 & $75 \pm \scriptstyle 200$ & 0.00\% & 0.00\% & $75 \pm \scriptstyle 239$ & 0.00\% & 0.00\% & $91 \pm \scriptstyle 249$ & 0.00\% & 0.00\% & $201 \pm \scriptstyle 381$ & 0.00\% & 0.00\% \\
        \midrule
        365 & 1 & $137 \pm \scriptstyle 121$ & 3.96\% & 2.97\% & $239 \pm \scriptstyle 280$ & 0.00\% & 0.00\% & $314 \pm \scriptstyle 322$ & 0.00\% & 0.00\% & $329 \pm \scriptstyle 364$ & 0.00\% & 0.00\% \\
        365 & 10 & $58 \pm \scriptstyle 132$ & 1.98\% & 0.00\% & $81 \pm \scriptstyle 225$ & 0.00\% & 0.00\% & $104 \pm \scriptstyle 243$ & 0.00\% & 0.00\% & $216 \pm \scriptstyle 388$ & 0.00\% & 0.00\% \\
        365 & 50 & $52 \pm \scriptstyle 140$ & 0.00\% & 0.00\% & $52 \pm \scriptstyle 190$ & 0.00\% & 0.00\% & $68 \pm \scriptstyle 203$ & 0.00\% & 0.00\% & $200 \pm \scriptstyle 381$ & 0.00\% & 0.00\% \\
        \midrule
        50 & 1 & $101 \pm \scriptstyle 133$ & 6.93\% & 5.94\% & $239 \pm \scriptstyle 284$ & 0.00\% & 0.00\% & $309 \pm \scriptstyle 312$ & 0.00\% & 0.00\% & $334 \pm \scriptstyle 380$ & 0.00\% & 0.00\% \\
        50 & 10 & $35 \pm \scriptstyle 54$ & 3.96\% & 1.98\% & $75 \pm \scriptstyle 223$ & 0.00\% & 0.00\% & $94 \pm \scriptstyle 230$ & 0.00\% & 0.00\% & $217 \pm \scriptstyle 391$ & 0.00\% & 0.00\% \\
        50 & 50 & $37 \pm \scriptstyle 91$ & 0.00\% & 0.00\% & $45 \pm \scriptstyle 182$ & 0.00\% & 0.00\% & $60 \pm \scriptstyle 191$ & 0.00\% & 0.00\% & $196 \pm \scriptstyle 370$ & 0.00\% & 0.00\% \\
        \midrule
        10 & 1 & $89 \pm \scriptstyle 180$ & 6.93\% & 6.93\% & $258 \pm \scriptstyle 296$ & 0.00\% & 0.00\% & $327 \pm \scriptstyle 314$ & 0.00\% & 0.00\% & $384 \pm \scriptstyle 469$ & 0.00\% & 0.00\% \\
        10 & 10 & $9 \pm \scriptstyle 126$ & 4.95\% & 4.95\% & $83 \pm \scriptstyle 244$ & 0.00\% & 0.00\% & $90 \pm \scriptstyle 222$ & 0.00\% & 0.00\% & $218 \pm \scriptstyle 391$ & 0.00\% & 0.00\% \\
        10 & 50 & $25 \pm \scriptstyle 28$ & 1.98\% & 1.98\% & $47 \pm \scriptstyle 191$ & 0.00\% & 0.00\% & $61 \pm \scriptstyle 195$ & 0.00\% & 0.00\% & $195 \pm \scriptstyle 368$ & 0.00\% & 0.00\% \\
        \bottomrule
    \end{tabular}
    }
    \label{tab:app-fmow-natural}
\end{table*}

\begin{table*}[!ht]
    \caption{
    Results for \textbf{set prediction with a temporal shift on Naval propulsion} (\autoref{subsec:exp-sets}). We provide monitoring results of the miscoverage rate for $\epsilon=0.1, \delta=0.1$ and different combinations of the two key tracking parameters, sliding window size $S$ and batch size $B$. The delay quantity $\bar{\tau}(\psi)$ denotes the mean and std. deviation of detection delays $\tau(\psi) - \tau_*(\psi)$ across repeated trials ($R=50$), whereas $\%FP > 0$ and $\%FP > \delta$ denote the number of thresholds $\psi \in \gPsi$ (in \%) whose false alarm rate is non-zero or exceeds the desired rate $\delta$, respectively.
    }
    \resizebox{\linewidth}{!}{
    \begin{tabular}{llrrrrrrrrrrrr}
        \toprule
        \multicolumn{2}{c}{\textbf{Params}} & \multicolumn{3}{c}{\textbf{Running Risk $\hat{\gR}_r(\psi)$}} & \multicolumn{3}{c}{\textbf{Wealth Process $M_t(\psi)$}} & \multicolumn{3}{c}{\textbf{Wealth Process $M^{SUM}_t(\psi)$}} & \multicolumn{3}{c}{\textbf{Wealth Process $M^{EB}_t(\psi)$}} \\
        \cmidrule(lr){3-5} \cmidrule(lr){6-8} \cmidrule(lr){9-11} \cmidrule(lr){12-14}
        \textbf{$S$} & \textbf{$B$} & Delay $\bar{\tau}(\psi)$ & \%FP $> 0$ & \%FP $> \delta$ & Delay $\bar{\tau}(\psi)$ & \%FP $> 0$ & \%FP $> \delta$ & Delay $\bar{\tau}(\psi)$ & \%FP $> 0$ & \%FP $> \delta$ & Delay $\bar{\tau}(\psi)$ & \%FP $> 0$ & \%FP $> \delta$ \\
        \midrule
        None & 1 & $1373 \pm \scriptstyle 952$ & 0.00\% & 0.00\% & $1416 \pm \scriptstyle 957$ & 0.00\% & 0.00\% & $1469 \pm \scriptstyle 971$ & 0.00\% & 0.00\% & $3527 \pm \scriptstyle 2193$ & 0.00\% & 0.00\% \\
        None & 10 & $1364 \pm \scriptstyle 963$ & 0.00\% & 0.00\% & $1401 \pm \scriptstyle 969$ & 0.00\% & 0.00\% & $1450 \pm \scriptstyle 983$ & 0.00\% & 0.00\% & $3216 \pm \scriptstyle 1969$ & 0.00\% & 0.00\% \\
        None & 50 & $1361 \pm \scriptstyle 965$ & 0.00\% & 0.00\% & $1400 \pm \scriptstyle 970$ & 0.00\% & 0.00\% & $1449 \pm \scriptstyle 984$ & 0.00\% & 0.00\% & $3169 \pm \scriptstyle 1946$ & 0.00\% & 0.00\% \\
        \midrule
        200 & 1 & $293 \pm \scriptstyle 453$ & 0.00\% & 0.00\% & $419 \pm \scriptstyle 540$ & 0.00\% & 0.00\% & $492 \pm \scriptstyle 660$ & 0.00\% & 0.00\% & $1178 \pm \scriptstyle 1076$ & 0.00\% & 0.00\% \\
        200 & 10 & $305 \pm \scriptstyle 491$ & 0.00\% & 0.00\% & $349 \pm \scriptstyle 503$ & 0.00\% & 0.00\% & $375 \pm \scriptstyle 507$ & 0.00\% & 0.00\% & $1104 \pm \scriptstyle 992$ & 0.00\% & 0.00\% \\
        200 & 50 & $308 \pm \scriptstyle 492$ & 0.00\% & 0.00\% & $325 \pm \scriptstyle 498$ & 0.00\% & 0.00\% & $346 \pm \scriptstyle 501$ & 0.00\% & 0.00\% & $1074 \pm \scriptstyle 990$ & 0.00\% & 0.00\% \\
        \midrule
        50 & 1 & $146 \pm \scriptstyle 297$ & 54.46\% & 8.91\% & $394 \pm \scriptstyle 514$ & 0.00\% & 0.00\% & $438 \pm \scriptstyle 537$ & 0.00\% & 0.00\% & $1077 \pm \scriptstyle 1059$ & 0.00\% & 0.00\% \\
        50 & 10 & $204 \pm \scriptstyle 430$ & 0.00\% & 0.00\% & $318 \pm \scriptstyle 505$ & 0.00\% & 0.00\% & $342 \pm \scriptstyle 507$ & 0.00\% & 0.00\% & $1011 \pm \scriptstyle 973$ & 0.00\% & 0.00\% \\
        50 & 50 & $232 \pm \scriptstyle 482$ & 0.00\% & 0.00\% & $284 \pm \scriptstyle 498$ & 0.00\% & 0.00\% & $312 \pm \scriptstyle 501$ & 0.00\% & 0.00\% & $915 \pm \scriptstyle 866$ & 0.00\% & 0.00\% \\
        \midrule
        10 & 1 & $58 \pm \scriptstyle 147$ & 81.19\% & 60.40\% & $416 \pm \scriptstyle 544$ & 0.00\% & 0.00\% & $451 \pm \scriptstyle 566$ & 0.00\% & 0.00\% & $1024 \pm \scriptstyle 1055$ & 0.00\% & 0.00\% \\
        10 & 10 & $99 \pm \scriptstyle 186$ & 1.98\% & 0.00\% & $330 \pm \scriptstyle 511$ & 0.00\% & 0.00\% & $336 \pm \scriptstyle 507$ & 0.00\% & 0.00\% & $985 \pm \scriptstyle 968$ & 0.00\% & 0.00\% \\
        10 & 50 & $191 \pm \scriptstyle 463$ & 0.00\% & 0.00\% & $284 \pm \scriptstyle 500$ & 0.00\% & 0.00\% & $311 \pm \scriptstyle 502$ & 0.00\% & 0.00\% & $888 \pm \scriptstyle 861$ & 0.00\% & 0.00\% \\
        \bottomrule
    \end{tabular}
    }
    \label{tab:app-uci-natural}
\end{table*}

\end{document}